\documentclass[journal,print]{IEEEtran} %\documentclass[journal,twoside]{ieeecolor} % web
\usepackage{cite}
\usepackage{amsmath,amssymb,amsfonts}
\usepackage{algorithmic}
\usepackage{graphicx}
\usepackage{subcaption}
\usepackage{multirow}
\usepackage{algorithm,algorithmic}
\usepackage{threeparttable}
\usepackage{float}
\usepackage{balance}
\usepackage{hyperref}
\hypersetup{hidelinks=true}
\usepackage{textcomp}
\def\BibTeX{{\rm B\kern-.05em{\sc i\kern-.025em b}\kern-.08em
    T\kern-.1667em\lower.7ex\hbox{E}\kern-.125emX}}
\begin{document}
\title{Logit-Level Uncertainty Quantification in Vision-Language Models for Histopathology Image Analysis}
% \author{First A. Author, \IEEEmembership{Fellow, IEEE}, Second B. Test, and Third C. Author Jr., \IEEEmembership{Member, IEEE}}

\author{Betul Yurdem, Ferhat Ozgur Catak, Murat Kuzlu, and Mehmet Kemal Gullu
%\thanks{This paragraph of the first footnote will contain the date on which you submitted your paper for review. It will also contain support information, including sponsor and financial support acknowledgment. For example, ``This work was supported in part by the U.S. Department of Commerce under Grant 123456.'' }
\thanks{Betul Yurdem and Mehmet Kemal Gullu are with the Department of Electrical and Electronics Engineering, Izmir Bakircay University, 35665 Izmir, Turkey (e-mail: betul.yurdem@bakircay.edu.tr; kemal.gullu@bakircay.edu.tr).}
\thanks{Ferhat Ozgur Catak is with the Department of Electrical Engineering and Computer Science, University of Stavanger, 4021 Stavanger, Norway (e-mail: f.ozgur.catak@uis.no).}
\thanks{Murat Kuzlu is with the Batten College of Engineering and Technology, Old Dominion University, Norfolk, VA 23529, USA (e-mail: mkuzlu@odu.edu).}}

\maketitle

\begin{abstract}
Vision-Language Models (VLMs) with their multimodal capabilities have demonstrated remarkable success in almost all domains, including education, transportation, healthcare, energy, finance, law, and retail. Nevertheless, the utilization of VLMs in healthcare applications raises crucial concerns due to the sensitivity of large-scale medical data and the trustworthiness of these models (reliability, transparency, and security). This study proposes a logit-level uncertainty quantification (UQ) framework for histopathology image analysis using VLMs to deal with these concerns. UQ is evaluated for three VLMs using metrics derived from temperature-controlled output logits. The proposed framework demonstrates a critical separation in uncertainty behavior. While VLMs show high stochastic sensitivity (cosine similarity (CS) $<0.71$ and $<0.84$, Jensen-Shannon divergence (JS) $<0.57$ and $<0.38$, and Kullback-Leibler divergence (KL) $<0.55$ and $<0.35$, respectively for mean values of VILA-M3-8B and LLaVA-Med v1.5), near-maximal temperature impacts ($\Delta_T \approx 1.00$), and displaying abrupt uncertainty transitions, particularly for complex diagnostic prompts. In contrast, the pathology-specific PRISM model maintains near-deterministic behavior (mean CS $>0.90$, JS $<0.10$, KL $<0.09$) and significantly minimal temperature effects across all prompt complexities. These findings emphasize the importance of logit-level uncertainty quantification to evaluate trustworthiness in histopathology applications utilizing VLMs.

\end{abstract}

\begin{IEEEkeywords}Vision Language Models (VLMs), Uncertainty Quantification, Histopathology, Trustworthy AI 
% Enter keywords or phrases in alphabetical order, separated by commas. Using the IEEE Thesaurus can help you find the best standardized keywords to fit your article. Use the thesaurus access request form for free access to the IEEE Thesaurus: \underline{https://www.ieee.org/publications/services/thesaurus-acce}\\
% \underline{ss-page.com.}
\end{IEEEkeywords}

\section{Introduction}
\label{sec:introduction}
In recent years, the Artificial Intelligence (AI) landscape has undergone a significant transformation, driven by the introduction of powerful Generative AI models, particularly Large Language Models (LLMs) and Vision Language Models (VLMs). LLMs such as GPT series, LLaMA, Gemini, and Claude enable machines to understand and generate text, while VLMs such as CLIP, BLIP, and LLaVA understand images along with text to analyze them, and answer visual questions \cite{bordes2024introduction}. These models have been applied to a variety of domains, including education, transportation, healthcare, finance, law, and retail, with remarkable success. Among these domains, healthcare is more critical due to the direct effect of decisions on human lives, and requires decisions to be more accurate, safe, and trustworthy than in any other domain. In addition, medical data processing is more complex, highly sensitive, and often scarce. VLMs have shown more promising performance in healthcare applications by enabling multimodal (images and text) reasoning over clinical data. However, many challenges need to be addressed to improve the trustworthiness of those models in terms of reliability, transparency, and security, particularly in high-stakes medical domains such as diagnosis to minimize risk, and support human–AI collaboration.

In the literature, a variety of studies have been published on AI/ML models focusing on cancer diagnosis, histopathology image analysis, and interpretation \cite{araujo2017classification, saxena2020machine, liu2022classification}, which have been more popular in healthcare applications in recent years, due to their remarkable success as a complementary tool for faster, more comprehensive reporting and efficient healthcare providers. 
However, most existing models are limited by their capabilities, datasets, evaluation methods, and performance due to their unimodality and patient data privacy. The study \cite{guo2024histgen} proposes a Multiple Instance Learning (MIL)-based framework for histopathology report generation, called HistGen. It also provides a high-quality dataset of Whole Slide Image (WSI) reports, including around 7,800 WSI-report pairs. 
% The HistGen framework aims to enable automated WSI report generation and diagnostic reports from local and global feature encoding. 
According to the results, the HistGen framework offers a better performance than state-of-the-art (SOTA) methods on WSI report generation on cancer subtyping and survival analysis tasks. A quantitative and explainable analysis study \cite{nguyen2024towards} compares histopathology embeddings generated from a pre-trained VLM with a word-of-interest pool, selecting the most similar words to generate a text-based image embedding. Another study \cite{bui2025lifelong} proposes the ADaFGrad framework to enhance lifelong learning for histopathology WSI analysis. A strong performance is achieved with fewer training sessions with this framework, which includes Online Vision-Language Adaptation (OVLA), allowing the selection of the equivalent of patch features of a slide from a predefined set of prototype texts, and Past-to-Present Gradient Distillation (PPGD), allowing preserving memory by comparing logit gradients in past and current iterations when the number of tasks increases. For various histopathology-related tasks, the authors in \cite{gilal2025pathvlm} provide a comprehensive evaluation benchmark of VLMs, including LLaVA, Qwen-VL, Qwen2-VL, InternVL, Phi3, Llama3, MOLMO, and XComposer, on the PathMMU dataset. The study also compares VLMs in terms of the relationship between model size and performance, and the results indicate that larger models, demanding higher computing, typically achieve higher accuracy, e.g., Qwen2-VL-72B-Instruct's average score is 63.97\%, due to a better multimodal representation learning and instructions, multi-step reasoning, and more knowledge. Recent studies emphasize the critical role of UQ for evaluating VLMs. For instance, a study using conformal prediction for uncertainty quantification in VLMs demonstrates that uncertainty is often more pivotal than assessing raw response accuracy \cite{kostumov2024uncertainty}. In this direction, the ViLU framework \cite{lafon2025vilu} was designed specifically for VLM uncertainty quantification by predicting errors without requiring access to the internal parameters of the VLM by leveraging a weighted binary cross-entropy loss. In the literature, various metrics are utilized for VLM uncertainty quantification, particularly in medical applications, e.g., cosine similarity (CS) was widely utilized to produce uncertainty-aware similarity values within a contrastive learning framework for medical image segmentation \cite{pan2025dusss}.  However, as highlighted in \cite{pan2025evivlm}, employing a combination of complementary metrics provides more reliable and meaningful insights into uncertainty quantification than relying on a single metric. Consequently, divergence-based metrics have gained traction. Kullback–Leibler (KL) divergence is effectively applied to detect hallucinations and support self-training in LLMs for math reasoning, and VLM for attribution binding tasks \cite{yangunderstanding}. Furthermore, the Jensen-Shannon (JS) divergence, a symmetric and more stable version of the KL divergence, is increasingly preferred in several studies \cite{venkataramanan2025probabilistic, lin2025diq, imam2025t3} for its effectiveness in comparing distributional differences in VLM uncertainty quantification evaluation. 

Although there are significant advancements in GAI models with advanced computing technologies, a limited uncertainty quantification study specifically targeting histopathology-related VLMs was identified in the literature. To address this critical deficiency, this paper presents a method involving a systematic and logit-level uncertainty quantification analysis of three prominent VLMs, VILA-M3-8B, LLaVA-Med v1.5, and PRISM, using cross-entropy (CE), Kullback-Leibler divergence, JS divergence, and mean absolute error (MAE) as evaluation metrics. Balancing computational power efficiency while comprehensively assessing the chosen VLMs is another challenge; therefore, a representative set of 100 histopathology patches is curated to sufficiently encompass the embedding spaces. To assess VLM uncertainty quantification, 11 temperature values ranging from 0.0 to 1.0 were applied to VLMs. Furthermore, to assess the robustness of the model and stochastic variability, 30 iterations are performed for each combination of histopathologic images and three different diagnostic prompts as input. Finally, metrics are computed directly from the saved output logits, providing a rigorous uncertainty quantification of histopathology-related VLMs.

\section{System Overview}

\begin{figure*}[h]
    \centering
    \includegraphics[width=0.95\linewidth]{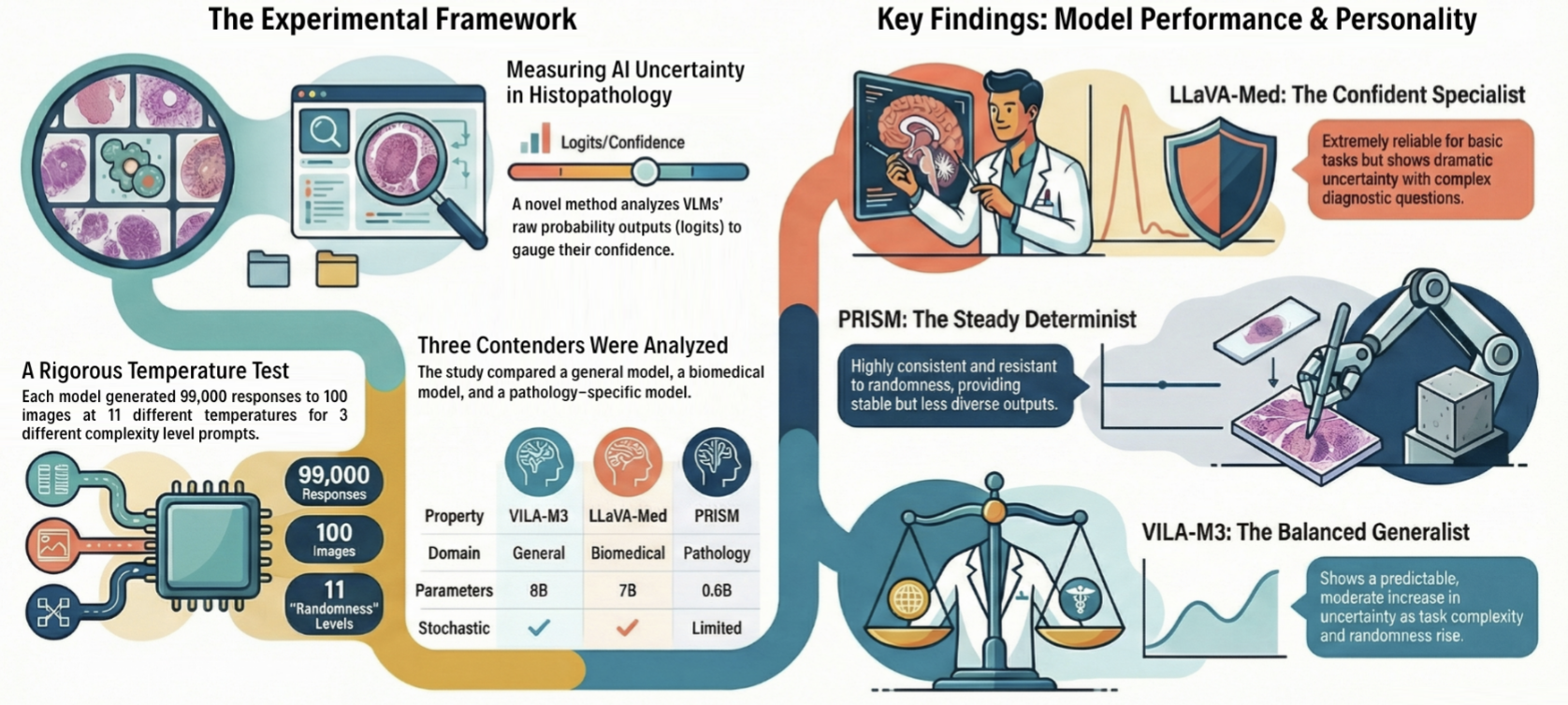}
    \caption{Diagram with key highlights from the proposed logit-level uncertainty quantification framework.} 
\label{fig:system_overview}%\vspace{-3mm}
\end{figure*}

The system is designed to be able to perform for three heterogeneous models, i.e, VILA-M3-8B \cite{nath2024vila}, LLaVA-Med v1.5 \cite{li2023llavamed}, and PRISM \cite{shaikovski2024prism}, despite their architectural differences (Table~\ref{tab:model_comparison}). The pipeline combines embedding-space visualization, image embedding extraction, temperature-scaled autoregressive decoding, logit capture, and the computation of pairwise uncertainty metrics. The general description diagram of the system is given in Figure~\ref{fig:system_overview}, which was generated with the help of AI, manually edited, and improved by the authors. %The overview of the system is given in Figure~\ref{fig:system_overview}.

% Optional: Add a table summarizing the models
\begin{table}[h]
\centering
\caption{Comparison of Vision-Language Models}
\label{tab:model_comparison}
\begin{tabular}{lccc}
\hline
\textbf{Property} & \textbf{VILA-M3} & \textbf{LLaVA-Med} & \textbf{PRISM} \\
\hline
Parameters & 8B & 7B & 0.6B \\
Domain & General & Biomedical & Pathology \\
Vision Encoder & CLIP & CLIP & Virchow \\
Base LLM & Llama-3 & Mistral & Custom \\
KV-Cache & \checkmark & \checkmark & $\times$ \\
Stochastic & \checkmark & \checkmark & Limited \\
\hline
\end{tabular}
\end{table}

% \begin{figure*}[h]
%     \centering
%     \includegraphics[width=1.0\linewidth]{imgs/system_overview.pdf}
%     \caption{Overview of the proposed logit-level uncertainty quantification framework.} 
%     \label{fig:system_overview}%\vspace{-3mm}
% \end{figure*}

\subsection{High-Level Description}

% Given a histopathology patch $I$ and a diagnostic prompt $p$, the system executes a multi-stage workflow. First, each model generates an image embedding that reflects how its visual encoder interprets the tissue morphology. The extracted embeddings are projected into a two-dimensional space using t-distributed stochastic neighbor embedding (t-SNE) to reveal structural differences in the models' visual feature representations. Next, the model generates textual output in an autoregressive manner while the underlying logits at every generation step are captured. This decoding process is repeated $N=30$ times for each of eleven temperature values varying in 0.1 increments from $T=0.0$ (deterministic greedy decoding) to $T=1.0$ (maximal sampling entropy). Finally, these recorded results form the basis for a detailed pairwise comparison of the resulting logit sequences using several complementary divergence and similarity measures.
The proposed system executes a multi-stage pipeline by giving a histopathology patch $I$ and a clinical diagnostic question $p$. Initially, the visual encoder of each VLM generates an image embedding, which reflects the interpretation of tissue morphology. To reveal structural differences across the visual feature representation of models, the extracted embeddings are mapped into a two-dimensional space using t-distributed stochastic neighbor embedding (t-SNE). After that, each model generates textual diagnostic outputs autoregressively while the underlying logits at every decoding step are meticulously saved. This decoding phase is repeated $N=30$ times for each of 11 temperatures ranging from $T=0.0$ (deterministic greedy decoding) to $T=1.0$ (maximal sampling entropy) in increments of 0.1. For a detailed pairwise comparison of these logit sequences, several divergence and similarity metrics are computed.

% This multi-level approach allows us to analyze uncertainty not only in the generated textual tokens, but also within the \emph{latent probability geometry} from which the outputs arise. By operating directly in logit and embedding space, the proposed system captures subtle but diagnostically relevant model behaviors that would be obscured if one only analyzed terminal text.
A comprehensive uncertainty quantification analysis at the token level, along with the hidden probability geometry derived from outputs, is enabled. The proposed framework detects subtle but diagnostically relevant model behaviors through operating in both the logit and the embedding spaces.

\subsection{Embedding-Space Characterization}

%While logits provide local, token-level uncertainty analysis, embeddings allow to examine global structure in the model’s visual representations. To reduce computational power, the test dataset is formed by selecting fewer images, 100, to cover the embedding space of each VLM. For each VLM, embeddings $\mathcal{E}_v$ of all images in the tested dataset are collected:
Whereas logits enable local and token-level uncertainty quantification, embeddings facilitate examination of the global structure within the visual representation space of each VLM. To mitigate computational demands and ensure comprehensive coverage, the evaluation dataset is constructed by selecting a reduced yet representative subset of 100 histopathology images to cover all of the embedding spaces of each VLM. For every VLM under study, visual embeddings $\mathcal{E}_v$ of all images in the evaluated dataset are aggregated as follows:
%~\eqref{eq:eq1}.

\begin{equation}
\mathcal{E}_v = \left\{ E_v(I_1), \dots, E_v(I_{i}) \right\}.
%\label{eq:eq1}
\end{equation}

Then, t-SNE is applied to map each of these embeddings into a 2D space:

\begin{equation}
\mathbf{u}_i = \mathrm{tSNE}(E_v(I_i)) \in \mathbb{R}^2
\end{equation}

Interpretable clusters are obtained from the resulting visualizations, which show separations among image types, space alignments, and model-specific specializations. A detailed understanding of VLM behavior is ensured through the concretion of \emph{local} (logit-level) and \emph{global} (embedding-level) uncertainty analysis in this proposed framework.

\subsection{Image Embedding Extraction}

To reflect the training methodology of selected VLMs, an image encoder $E_v(\cdot)$ is leveraged. The CLIP-style transformer, as an encoder of VILA-M3 and LLaVA-Med, generates a sequence of visual tokens as follows:%The selected VLM utilizes an image encoder $E_v(\cdot)$, reflecting their training methodology. VILA-M3 and LLaVA-Med rely on a CLIP-style transformer for their encoder to generate a sequence of visual tokens as follows:

\begin{equation}
E_v(I) = [\mathbf{v}_1, \ldots, \mathbf{v}_n], \mathbf{v}_n \in \mathbb{R}^{d_v}
\end{equation}
% where $\mathbf{v}_n \in \mathbb{R}^{d_v}$. 
Furthermore, PRISM works with Virchow, designed as a pathology-specific encoder. The encoder output is computed by concatenating a class of tokens with mean pooling:
\begin{equation}
\mathbf{e} = [\mathbf{e}_{\mathrm{cls}},~ \mathrm{MeanPool}(\mathbf{e}_{\mathrm{patch}})] \in \mathbb{R}^{2560}
\end{equation}

These visual embeddings make geometric analyses, such as model specialization and cluster division, more practical.

% Each evaluated VLM provides an image encoder $E_v(\cdot)$ tailored to its training regime. For VILA-M3 and LLaVA-Med, this encoder is a CLIP-style transformer operating on image patches, producing a sequence of visual tokens:
% \begin{equation}
%     E_v(I) = [\mathbf{v}_1, \ldots, \mathbf{v}_n], \qquad
%     \mathbf{v}_i \in \mathbb{R}^{d_v}.
% \end{equation}
% PRISM, in contrast, employs a pathology-specific Virchow encoder designed to capture fine-grained histomorphological patterns. Its representation is constructed through class-token concatenation and mean pooling:
% \begin{equation}
% \mathbf{e}
% =
% [\mathbf{e}_{\mathrm{cls}},~
% \mathrm{MeanPool}(\mathbf{e}_{\mathrm{patch}})]
% \in \mathbb{R}^{2560}.
% \end{equation}
% 
% These embeddings, stored for all images, facilitate downstream geometric analyses such as cluster separation and model specialization.

\subsection{Temperature-Dependent Autoregressive Generation}

% To quantify stochasticity in model behavior, the language decoder is probed under controlled temperature perturbations. Given prompt $p$ and image features $V$, the decoder produces hidden states:
To rigidly quantify model behavior along with uncertainty quantification in diagnostic generation, the language decoder is evaluated with temperature perturbations. Given a diagnostic prompt $p$ and image features $V$, the decoder computes hidden states as follows:

\begin{equation}
    \mathbf{h}_t = f_{\theta}(y_{<t}, V, p),
\end{equation}
% followed by the corresponding logits:
which are subsequently mapped to the corresponding logits:
\begin{equation}
    \mathbf{z}_t = W_o \mathbf{h}_t \;\in\; \mathbb{R}^{|V|}.
\end{equation}

% Temperature scaling manipulates the softness of the output distribution:
Temperature $T$ is scaled to manipulate the softness of the predicted output distribution:
\begin{equation}
P_T(y_t \mid y_{<t},I,p)
=
\mathrm{softmax}\!\left(\frac{\mathbf{z}_t}{T}\right).
\end{equation}

% For $T=0$, the model behaves deterministically, producing identical outputs across repetitions. For $T>0$, randomness is introduced via nucleus sampling ($p=0.9$), enabling controlled exploration of the model’s probability surface. The complete logit sequence for each run:
% \begin{equation}
% Z^{(i)} = [\mathbf{z}^{(i)}_1, \dots, \mathbf{z}^{(i)}_{T_i}]
% \end{equation}
% is saved for further analysis.
When $T=0$, the decoding process operates deterministically and generates identical outputs across repetitions. For $T>0$, controlled stochasticity is introduced through nucleus sampling ($p=0.9$), which allows controlled exploration of the probability surface of the model. For each independent run $i$, the full sequence of logits, $Z^{(i)} = [\mathbf{z}^{(i)}_1, \dots, \mathbf{z}^{(i)}_{T_i}]$, is stored for further analysis.

\subsection{Logit Tensor Normalization and Pairwise Comparisons}

% Since autoregressive generation may stop at different lengths across runs, sequences are aligned to their minimum shared length $T_{\min}$. Each aligned logit tensor is therefore represented as:
Since the autoregressive generation may result in different lengths across repeated runs, sequences are aligned to their minimum shared length $T_{\min}$. Each aligned logit tensor is uniformly shaped as:
\begin{equation}
Z^{(i)} \in \mathbb{R}^{T_{\min} \times |V|}.
\end{equation}

% For each temperature, the proposed system computes all $\binom{N}{2}=435$, where $N$ is the number of repetitions and equals 30, pairwise comparisons across four uncertainty metrics.
For each of four complementary uncertainty metrics, the proposed framework computes all possible $\binom{N}{2}=435$ pairwise comparisons, where $N$ is the number of repetitions and equals 30.

\paragraph{Cosine Similarity (CS)} This metric measures the angular alignment between corresponding logit vectors, providing an understanding of directional consistency.

%Measures angular alignment:
\begin{equation}
\mathrm{CS}(Z^{(i)},Z^{(j)})
=
\frac{1}{T_{\min}}
\sum_{t=1}^{T_{\min}}
\frac{
\mathbf{z}^{(i)}_t \cdot \mathbf{z}^{(j)}_t
}{
\|\mathbf{z}^{(i)}_t\|
\|\mathbf{z}^{(j)}_t\|
}
\end{equation}

\paragraph{Kullback–Leibler (KL) Divergence} The directional probability distribution mismatch is quantified asymmetrically.%Quantifies directional distribution mismatch.
\begin{equation}
\mathrm{KL}(Z^{(i)}\|Z^{(j)})
= \frac{1}{T_{\min}}
\sum_{t=1}^{T_{\min}}
\sum_{v=1}^{|V|}
P^{(i)}_{T,v}
\log
\frac{P^{(i)}_{T,v}}{P^{(j)}_{T,v}}
\end{equation}
% where \( P^{(i)}_{t,v} = \mathrm{softmax}(z^{(i)}_{t,v})\).
\paragraph{Jensen-Shannon (JS) Divergence} This symmetric metric, which is limited to the range $[0, \log 2]$ and is a version of KL divergence, measures distributional uncertainty as follows:%Provides a symmetric, bounded measure of distributional separation.
\begin{equation}
\mathrm{JS}(Z^{(i)}, Z^{(j)}) =
\frac{1}{2}
\mathrm{KL}(P^{(i)}_{T,v} \| M_{T,v}) +
\frac{1}{2}
\mathrm{KL}(P^{(j)}_{T,v} \| M_{T,v}),
\end{equation}
where \( M_{T,v} = \tfrac{1}{2}\left(P^{(i)}_{T,v} + P^{(j)}_{T,v}\right) \). \\

\paragraph{Mean Absolute Error (MAE)} The MAE is one of the widely used metrics in the AI/ML landscape. In this study, MAE is used to measure the raw logit-level variability between paired logit values.

\begin{equation}
\mathrm{MAE}(Z^{(i)},Z^{(j)}) =
\frac{1}{T_{\min}|V|}
\sum_{t=1}^{T_{\min}}
\sum_{v=1}^{|V|}
|z^{(i)}_{t,v} - z^{(j)}_{t,v}|
\end{equation}

These complementary metrics help to examine distinguished changes in the \emph{direction} in logit space (CS), the \emph{shape} of the probability distribution (KL and JS divergences), and the \emph{magnitude} of logits (MAE).

% These complementary metrics allow to distinguish changes in the \emph{shape} of the probability distribution (KL, JS), the \emph{magnitude} of logits (MAE), and the \emph{direction} in logit space (CS).

\subsection{Algorithmic Pipeline}

Algorithm~\ref{alg:pipeline} illustrates the entire pipeline for the uncertainty quantification (UQ) framework. This is model-agnostic and compatible with all VLMs.

\begin{algorithm}[h]
\caption{End-to-End UQ Framework Pipeline}%End-to-End Uncertainty Quantification Pipeline}
\label{alg:pipeline}
\begin{algorithmic}[1]
\REQUIRE VLM $M$, image $I$, prompt $p$, temperature set $\{T_k\}$, repeats $N$
\FOR{each model $M$}
    \STATE $\mathbf{e} \gets E_v(I)$  \COMMENT{Extract embedding}
    \FOR{each temperature $T_k$}
        \FOR{$i=1$ to $N$}
            \STATE Initialize $y=\langle\mathrm{BOS}\rangle$ and $Z^{(i)}=\emptyset$
            \WHILE{token limit not reached}
                \STATE $\mathbf{h}_t \gets f_{\theta}(y_{<t},I,p)$
                \STATE $\mathbf{z}_t \gets W_o \mathbf{h}_t$
                \STATE Append $\mathbf{z}_t$ to $Z^{(i)}$
                \STATE Sample $y_t$ from $\mathrm{softmax}(\mathbf{z}_t/T_k)$
                \IF{$y_t = \langle\mathrm{EOS}\rangle$}
                    \STATE \textbf{break}
                \ENDIF
            \ENDWHILE
            \STATE Save $Z^{(i)}$
        \ENDFOR
        \STATE Compute all pairwise metrics for $\{Z^{(i)}\}$
    \ENDFOR
\ENDFOR
\end{algorithmic}
\end{algorithm}

\subsection{Experimental Design}

\subsubsection{Dataset Configuration}

The experimental setup includes these components:
\begin{itemize}
    \item \textbf{Images}: 100 histopathology patches, selected from the ARCH dataset \cite{gamper2020multiple}.
    \item \textbf{Prompts}: 3 diagnostic complexity levels.
    \begin{enumerate}
        \item Basic cellular morphology assessment
        \item Intermediate tissue diagnosis with grading
        \item Advanced systematic quantitative analysis
    \end{enumerate}
    \item \textbf{Temperature sweep}: 11 uniformly sampled temperature values $(T \in \{0.0, 0.1, 0.2, \ldots, 1.0\})$.
    \item \textbf{Repetitions}: $N = 30$ per configuration.
    \item \textbf{Total generations}: $100 \times 3 \times 11 \times 30 = 99{,}000$ per model.
\end{itemize}

\subsubsection{Computational Efficiency} 

\textbf{Caching Strategy}: For VILA-M3 and LLaVA-Med, key-value (KV) caching is applied to accelerate the generation process:

\begin{equation}
\mathbf{h}_t = f(\mathbf{h}_{t-1}, \text{KV-cache}_{<t}, y_{t-1})
\label{eq:kv_cache}
\end{equation}

This reduces computational complexity from $\mathcal{O}(T^2)$ to $\mathcal{O}(T)$ for sequence generation.

%For VILA-M3 and LLaVA-Med, a key-value (KV) cache is implemented to accelerate generation:
%This reduces computational complexity from $\mathcal{O}(T^2)$ to $\mathcal{O}(T)$ for sequence generation.

\textbf{PRISM Deterministic Behavior}: Even with temperature scaling, this VLM shows the least stochastic variation and does not utilize such caching mechanisms. Thus, PRISM requires another sampling strategy, such as Gaussian noise injection ($\mathcal{N}(0, \sigma_T^2)$) and dropout-based perturbation.

%PRISM exhibits minimal stochastic variation even with temperature scaling, does not utilize such caching mechanisms, necessitating alternative sampling strategies including Gaussian noise injection ($\mathcal{N}(0, \sigma_T^2)$) and dropout-based perturbation.

\subsection{Key Innovations}

Novel contributions are offered to the literature with this framework.

\begin{enumerate}
  \item \textbf{Logit-level uncertainty quantification}: Unlike token-level diversity metrics, distributional uncertainty is directly captured in the continuous probability space. 
  \item \textbf{Multi-model comparative analysis}: For general-purpose, biomedical, and pathology-specific VLMs, a comprehensive and systematic assessment of uncertainty quantification is performed.
  \item \textbf{Temperature-dependent uncertainty characterization}: A rigid quantification of how varying levels of stochastic sampling randomness, controlled with temperature scaling, influence prediction confidence and model stability in generated outputs. 
  \item \textbf{Prompt complexity stratification}: An assessment of uncertainty as a function of diagnostic task complexity enables the evaluation of model robustness across a spectrum of clinically relevant prompt difficulties. 
\end{enumerate}

\begin{figure*}[t]
    \centering
    \begin{subfigure}{0.3\textwidth}
    \centering
    \includegraphics[width=\linewidth]{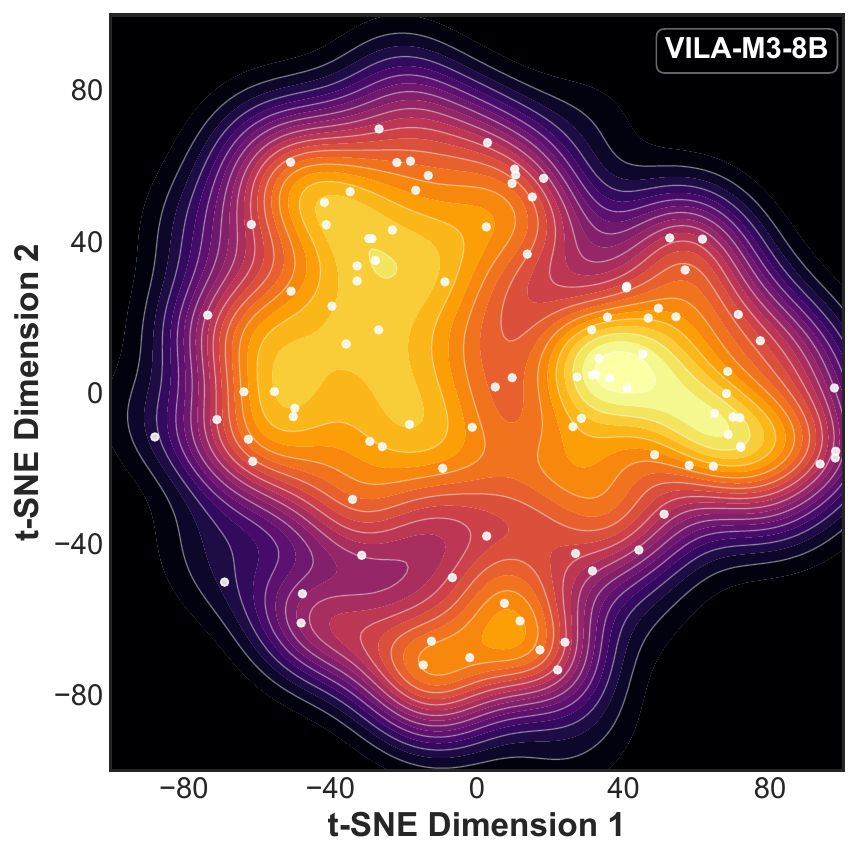}
    \caption{}
    \end{subfigure}
    \begin{subfigure}{0.3\textwidth}
    \centering
    \includegraphics[width=\linewidth]{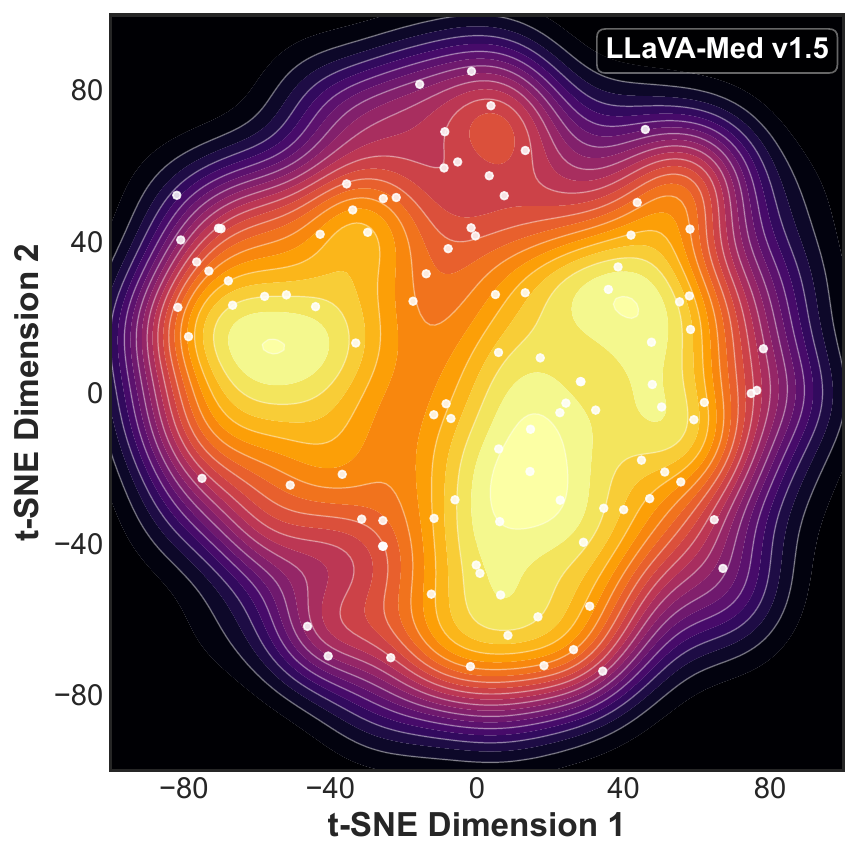}
    \caption{}
    \end{subfigure}
    \begin{subfigure}{0.3\textwidth}
    \centering
    \includegraphics[width=\linewidth]{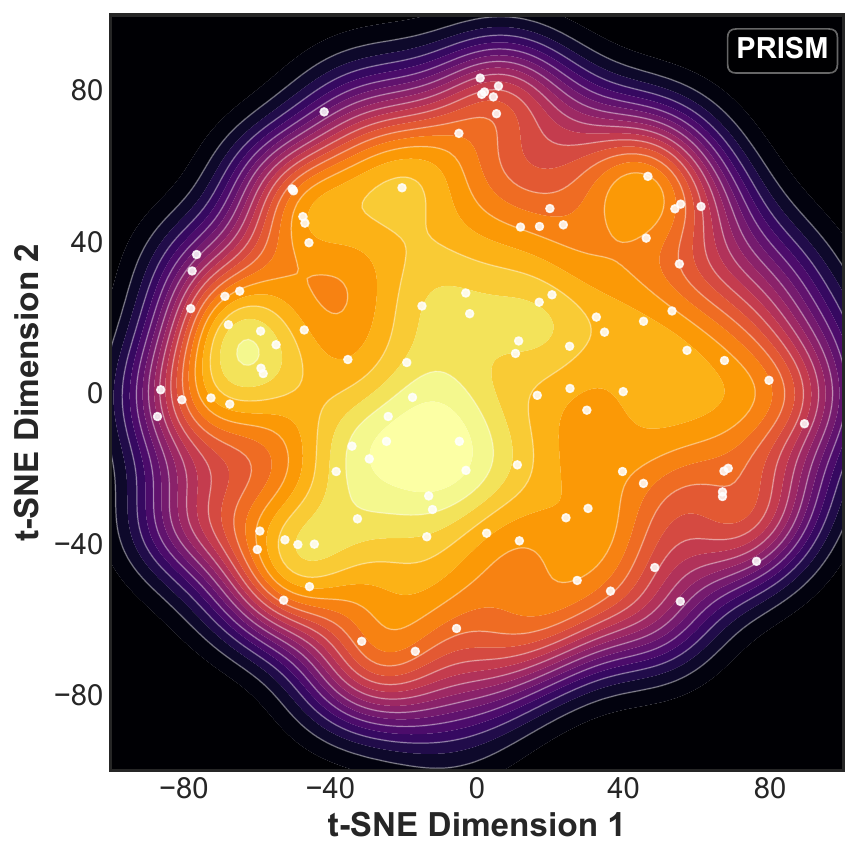}
    \caption{}
    \end{subfigure}    
    \caption{Embedding spaces of the evaluated VLMs and positions of the used histopathological patches.}
    \label{fig:embedding_spaces}
\end{figure*}

\subsection{Expected Outcomes}

The presented framework enables:
\begin{itemize}
    \item Identification of inputs inducing high VLM uncertainty,
    \item Comparative reliability assessment across VLM architectures,
    \item Temperature-optimal configurations for histopathology applications,
    \item Correlation analysis between uncertainty metrics and expert disagreement.
\end{itemize}

\section{Experimental Results}
\label{sec:results}

A comprehensive analysis of uncertainty quantification across three VLMs (VILA-M3-8B, LLaVA-Med v1.5, and PRISM) evaluated on 100 histopathology images with three prompt complexity levels is presented. All metrics are normalized within each model to enable cross-model comparison, yielding result values in the range $[0, 1]$.

\subsection{Embedding Space Analysis}

Since a comprehensive analysis is intended, the goal is to select images that will cover the entire embedding space of each model. However, due to limited computational power, a smaller dataset is curated to cover all spaces with fewer images. The embedding space visualizations for each VLM are given in Figure~\ref{fig:embedding_spaces}. The change in density across spaces shows the distribution of all images in the dataset, while the white dots mark 100 images selected to use in the study.

\subsection{Temperature-Dependent Uncertainty Behavior}

\subsubsection{Cosine Similarity Analysis}

\begin{figure*}[!t]
    \centering
    \begin{subfigure}{0.29\textwidth}
    \centering
    \includegraphics[width=\linewidth]{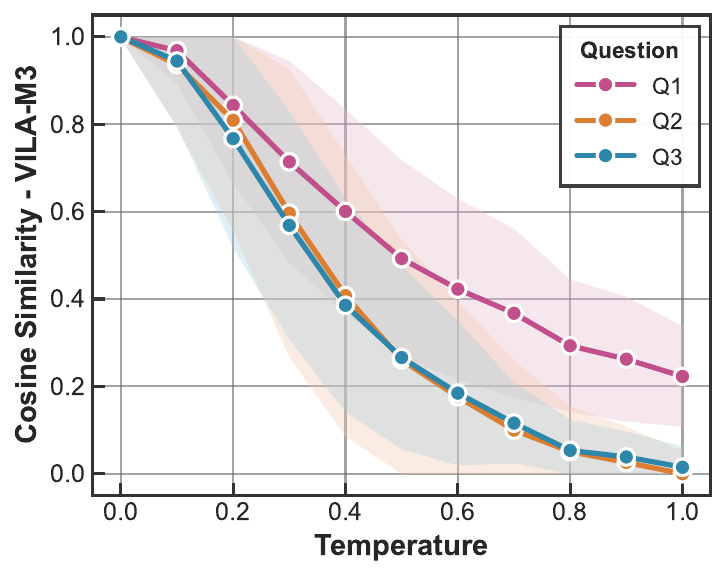}
    \caption{}
    \end{subfigure}\hfill
    \begin{subfigure}{0.29\textwidth}
    \centering
    \includegraphics[width=\linewidth]{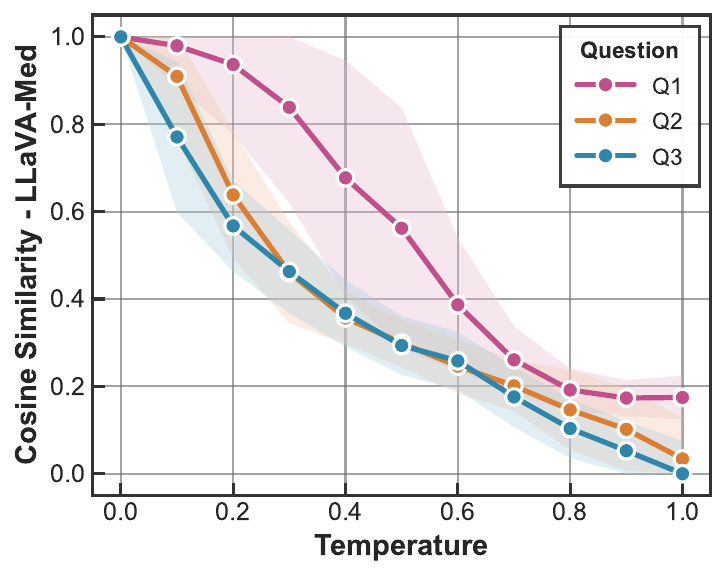}
    \caption{}
    \end{subfigure}\hfill
    \begin{subfigure}{0.29\textwidth}
    \centering
    \includegraphics[width=\linewidth]{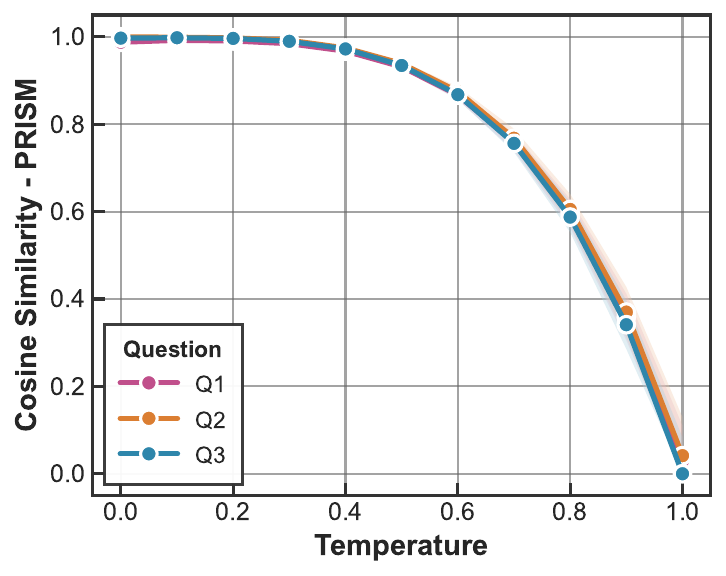}
    \caption{}
    \end{subfigure} \hfill   
    \caption{Normalized CS versus temperature for three VLMs across three question complexity levels is shown. Higher values indicate greater consistency between repeated iterations.} %PRISM exhibits minimal temperature sensitivity, while LLaVA-Med shows strong question-dependent degradation, particularly for complex prompts (Q3).}
    \label{fig:cosine_flat}
\end{figure*}

Figure~\ref{fig:cosine_flat} illustrates the normalized CS as a function of temperature across all models and question complexities. The main outcomes are listed below:

\begin{itemize}
    \item \textbf{VILA-M3} is the most stable one for all questions, among all VLMs. Its CS value is high ($>0.9$) for low temperatures ($T \leq 0.3$), and decreases monotonically. Depending on the question complexity, the CS is approximately 0.35--0.56 at $T=1.0$.
    
    \item \textbf{LLaVA-Med}'s performance strongly depends on the question:
    \begin{itemize}
        \item \textit{Q1 (Basic morphology):} While the temperature increases, the CS value gradually decreases from 1.0 to 0.04. Thus, the VLM has high stability for basic diagnostic tasks. %Gradual decline from 1.0 to 0.04 as temperature increases.
        \item \textit{Q2 (Intermediate diagnosis):} A similar but slightly faster decreasing pattern is observed for Q1. This highlights that the output of the model transitions from deterministic to stochastic decoding rapidly. %Similar pattern with slightly faster degradation.
        \item \textit{Q3 (Advanced quantitative):} The output consistency drops swiftly, and the CS approximate to zero ($\sim 0.02$) at the highest temperature. This results in this question causing the most severe degradation, and the model is extremely sensitive to advanced diagnostic tasks in sampling randomness. %Most severe degradation, approaching near-zero similarity ($\sim 0.02$) at high temperatures.
    \end{itemize}
    
    \item \textbf{PRISM} has high similarity ($>0.9$) at all temperatures up to $T=0.7$, and this confirms its deterministic structure. The CS values remain above 0.9 for all types of questions, even at $T=1.0$. This points to narrow stochastic variation despite temperature scaling. %maintains remarkably high similarity ($>0.9$) across all temperatures up to $T=0.7$, confirming its deterministic nature. Even at $T=1.0$, similarity remains above 0.9 for all question types, suggesting limited stochastic variation despite temperature scaling.
\end{itemize}

\subsubsection{Jensen-Shannon Divergence}

\begin{figure*}[!t]
    \centering
    \begin{subfigure}{0.29\textwidth}
    \centering
    \includegraphics[width=\linewidth]{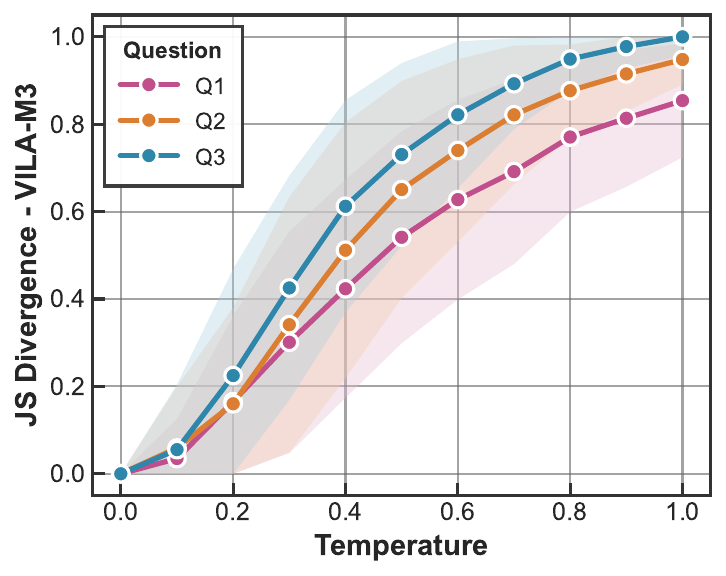}
    \caption{}
    \end{subfigure}\hfill
    \begin{subfigure}{0.29\textwidth}
    \centering
    \includegraphics[width=\linewidth]{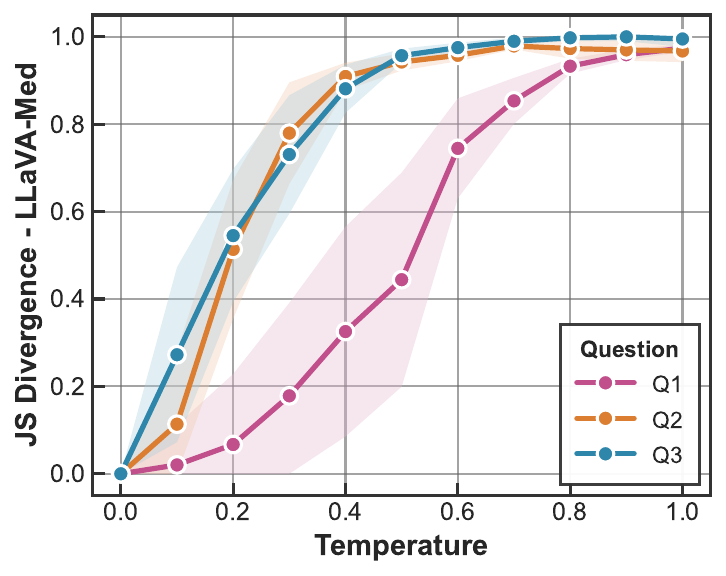}
    \caption{}
    \end{subfigure}\hfill
    \begin{subfigure}{0.29\textwidth}
    \centering
    \includegraphics[width=\linewidth]{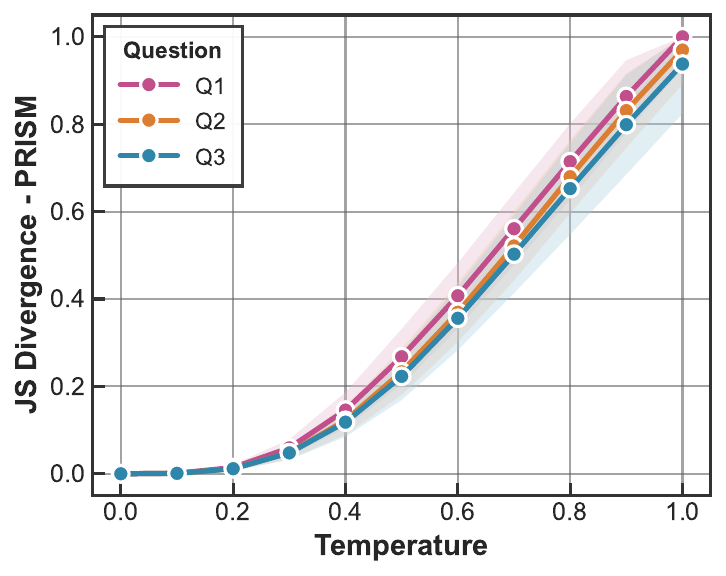}
    \caption{}
    \end{subfigure} \hfill   
    \caption{Normalized JS divergence versus temperature. Lower values indicate minimal uncertainty between repeated iterations.}% LLaVA-Med Q1 exhibits minimal uncertainty, while VILA-M3 shows consistent divergence growth with temperature.}
    \label{fig:js_divergence}
\end{figure*}

The JS divergence metric (Figure~\ref{fig:js_divergence}) measures the distributional uncertainty in the probability space. When this value increases, it indicates more variations over repeated iterations:%The JS divergence (Figure~\ref{fig:js_divergence}) quantifies distributional uncertainty in the probability space, with higher values indicating greater variability:

\begin{itemize}
    \item \textbf{VILA-M3} shows a smooth rise with temperature, reaching almost the maximum JS divergence value ($\sim 0.95$--1.0) at $T=1.0$ in all questions. In particular, the Q3 curve shows significant inconstancy in the middle of the temperature range ($T=0.4-0.7$). This highlights intricate interactions between sampling stochasticity and prompt complexity.    
    \item \textbf{LLaVA-Med} has significant uncertainty results at different complexities of the prompt levels. For Q1, a consistent low divergence ($<0.2$) is observed across the entire temperature range. This reflects robust confidence and stability. In the Q2 and Q3 results, the fastest rise from 0 to approximately 0.9 shows a continuous growth in uncertainty with temperature.
    % \begin{itemize}
    %     \item Q1: Minimal divergence ($<0.15$) across all temperatures, indicating high confidence.
    %     \item Q2: Abrupt transition at $T=0.7$, jumping from 0.15 to 0.93, then plateauing.
    %     \item Q3: Gradual increase from 0 to 0.68, suggesting continuous uncertainty growth.
    % \end{itemize}
    
    \item \textbf{PRISM} is almost zero ($<0.1$) over a large range of the temperature, while a slightly more increment at $T=1.0$ for Q1 is noticeable. This highlights the resistance of the VLM to temperature.%maintains near-zero JS divergence ($<0.1$) for most of the temperature range, with slight elevation ($\sim 0.3$) at $T=1.0$ for Q1. This confirms the model's resistance to temperature-induced stochasticity.
\end{itemize}

\subsubsection{Kullback-Leibler Divergence}

\begin{figure*}[!t]
    \centering
    \begin{subfigure}{0.29\textwidth}
    \centering
    \includegraphics[width=\linewidth]{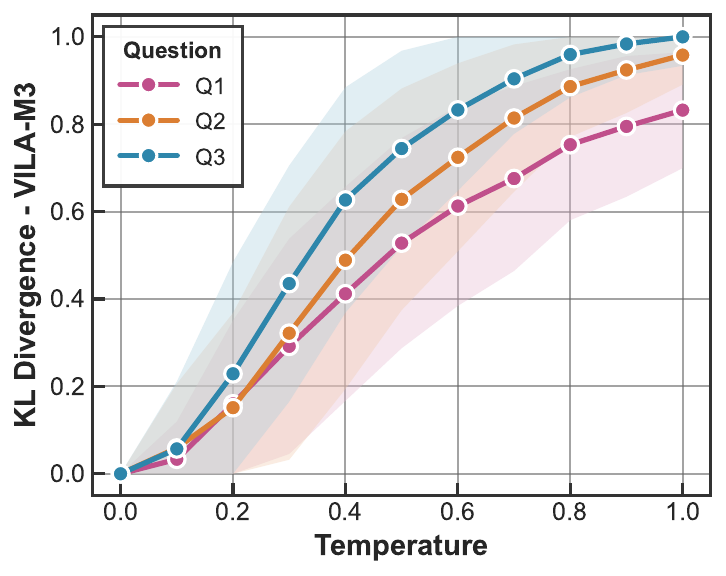}
    \caption{}
    \end{subfigure}\hfill
    \begin{subfigure}{0.29\textwidth}
    \centering
    \includegraphics[width=\linewidth]{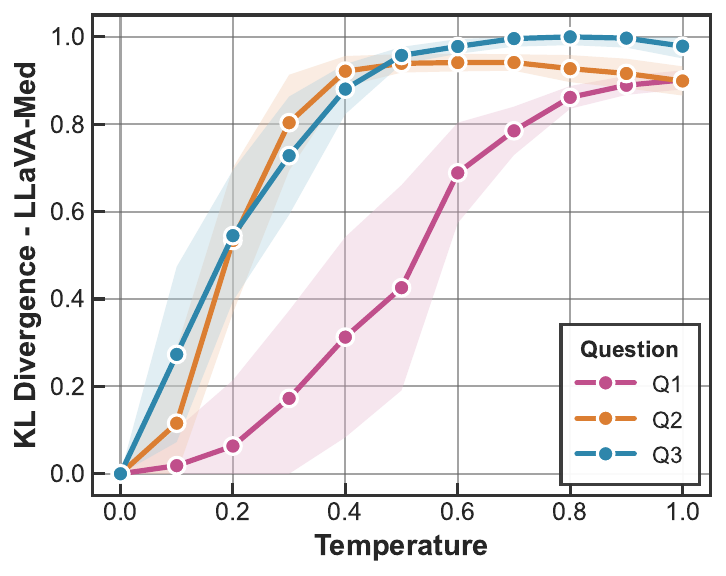}
    \caption{}
    \end{subfigure}\hfill
    \begin{subfigure}{0.29\textwidth}
    \centering
    \includegraphics[width=\linewidth]{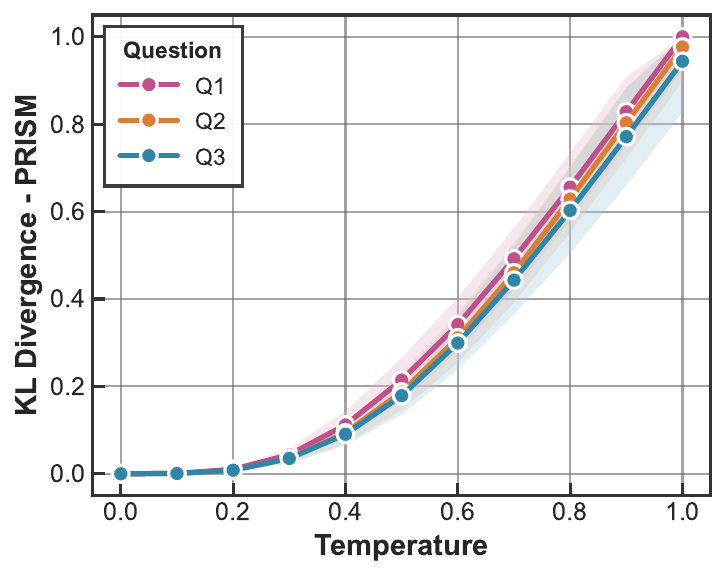}
    \caption{}
    \end{subfigure} \hfill   
    \caption{Normalized KL divergence versus temperature. Lower values indicate exceptional stability and highly reproducible probability distributions.} % LLaVA-Med Q1 exhibits exceptional stability (near-zero KL) across all temperatures, indicating highly reproducible probability distributions.} %$\text{KL}(P^{(i)} \| P^{(j)})$
    \label{fig:kl_ab}
\end{figure*}

Directional differences in probability distributions are revealed with asymmetric KL divergence (Figure~\ref{fig:kl_ab}):%The asymmetric KL divergence $\text{KL}(P^{(i)} \| P^{(j)})$ (Figure~\ref{fig:kl_ab}) reveals directional differences in probability distributions:

\begin{itemize}
    \item \textbf{VILA-M3} has a temperature-dependent behavior. The KL divergence grows and reaches saturation at higher temperatures ($\sim 0.8$--1.0), similar to JS divergence.  %exhibits temperature-dependent KL divergence growth similar to JS divergence, reaching saturation ($\sim 0.8$--1.0) at high temperatures. The Q3 curve shows characteristic oscillations in the $T=0.5-0.8$ range.
    
    \item \textbf{LLaVA-Med} is almost zero in KL divergence for Q1 through all temperatures. Thus, it behaves as a highly concentrated and stable probability distribution even under scaling. In contrast, Q2 and Q3 show slow increases at specific temperature thresholds between $T \approx 0.4$ and $T \approx 0.7$, reflecting abrupt shifts in uncertainty. %Q1 maintains near-zero KL divergence across the entire temperature spectrum, suggesting highly concentrated probability distributions even with temperature scaling. Q2 and Q3 show stepped increases at specific temperature thresholds ($T \approx 0.4$ and $T \approx 0.7$).
    
    \item \textbf{PRISM} aligns a pattern consistent with other metrics, with minimal divergence ($<0.1$) at $T \leq 0.7$ and moderate rises only at the higher temperatures.%follows a pattern consistent with other metrics: minimal divergence ($<0.1$) at $T \leq 0.7$, with moderate increases at extreme temperatures.
\end{itemize}

\subsubsection{Mean Absolute Error}

\begin{figure*}[!t]
    \centering
    \begin{subfigure}{0.29\textwidth}
    \centering
    \includegraphics[width=\linewidth]{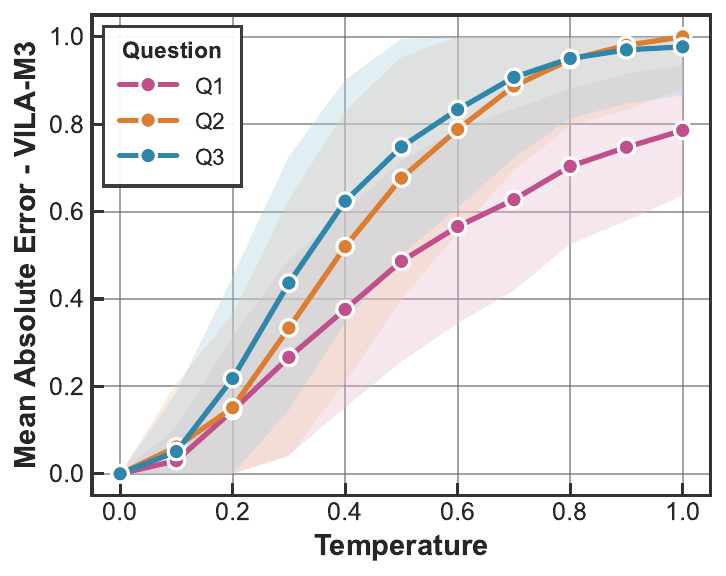}
    \caption{}
    \end{subfigure}\hfill
    \begin{subfigure}{0.29\textwidth}
    \centering
    \includegraphics[width=\linewidth]{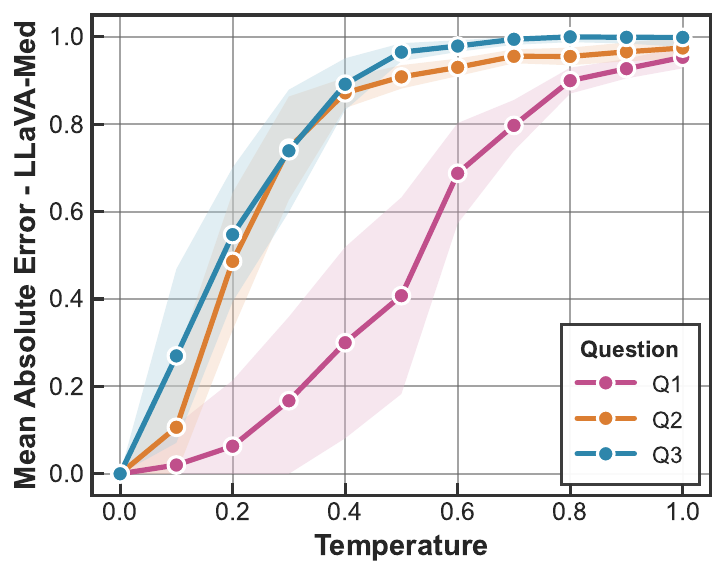}
    \caption{}
    \end{subfigure}\hfill
    \begin{subfigure}{0.29\textwidth}
    \centering
    \includegraphics[width=\linewidth]{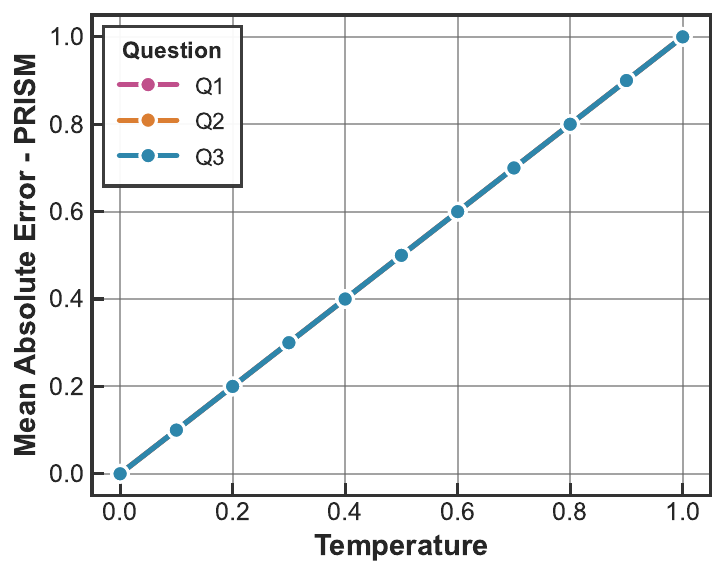}
    \caption{}
    \end{subfigure}  \hfill  
    \caption{Normalized MAE of logits versus temperature. VLMs show monotonic growth with convergence at higher temperatures.}% ($T \geq 0.8$).}%, indicating that temperature-induced randomness dominates model-specific behavior.}
    \label{fig:mae}
\end{figure*}

The MAE metric (Figure~\ref{fig:mae}) directly quantifies logit-level variability:%provides a direct measure of logit-level variability:

\begin{itemize}
    \item All evaluated models consistently demonstrate monotonic MAE growth with temperature increment, consistent with enhanced sampling diversity. %All models show monotonic MAE growth with temperature, consistent with increased sampling diversity.
    
    \item \textbf{VILA-M3} shows the slowest MAE increase for Q1, reaching only $\sim 0.786$ at $T=1.0$, whereas Q2 and Q3 results approach near-maximum values ($\sim 0.95$--1.0). %exhibits the slowest MAE growth for Q1, reaching only $\sim 0.61$ at $T=1.0$, while Q2 and Q3 approach near-maximum values ($\sim 0.8$--1.0).
    
    \item \textbf{LLaVA-Med} is the first one to reach saturation at $T \approx 0.4$ for Q2 and Q3, indicating to high variability. For Q1, a similar MAE trajectory indicates stable predictions.
    \item \textbf{PRISM} shows linear and similar increments for all questions. This means the question complexity does not affect the prediction stochasticity.
    % \item \textbf{LLaVA-Med and PRISM} show similar MAE trajectories for Q2 and Q3, converging to high values ($>0.8$) at $T=1.0$, indicating substantial logit-level variability.
    
    \item The convergence of all model-question combinations at high temperatures ($T \geq 0.8$) suggests that intense temperature scaling effectively overrides and dominates model- and prompt-specific characteristics.% The convergence of all model-question combinations at high temperatures ($T \geq 0.8$) suggests that extreme temperature scaling overwhelms model- and prompt-specific characteristics.
\end{itemize}


\subsection{Quantitative Summary}

Overall metric results for each VLM and diagnostic question pair are reported in Table~\ref{tab:model_question_metric}. For a high-level assessment of model stability, Table~\ref{tab:metric_summary} summarizes the aggregated statistics for every model–question combination across the entire temperature set. Specifically, the mean and standard deviation characterize all uncertainty levels, while the temperature effect $\Delta_{T}$ quantifies the normalized change from $T=0.0$ to $T=1.0$. %In Table~\ref{tab:model_question_metric}, overall metric results for each VLM and question are given. In summary, Table~\ref{tab:metric_summary} presents aggregate statistics for each model-question combination across the temperature spectrum. The mean and standard deviation characterize overall uncertainty levels, while the temperature effect $\Delta_{T}$ quantifies the normalized change from $T=0.0$ to $T=1.0$.

\begin{table*}[!t]
\centering
\caption{Summary of Normalized Uncertainty Metrics Across Temperature Range ($T \in [0, 1]$)}
\label{tab:model_question_metric}
\resizebox{\textwidth}{!}{%
\begin{tabular}{ccl|lllllllllll|}
\cline{4-14}
\multicolumn{1}{l}{\textbf{}} & \multicolumn{1}{l}{\textbf{}} & \textbf{} & \multicolumn{11}{c|}{\textbf{Temperatures}} \\ \hline
\multicolumn{1}{|c|}{\textbf{Metric}} & \multicolumn{1}{l|}{\textbf{Question}} & \textbf{Model} & \multicolumn{1}{l|}{\textbf{0.0}} & \multicolumn{1}{l|}{\textbf{0.1}} & \multicolumn{1}{l|}{\textbf{0.2}} & \multicolumn{1}{l|}{\textbf{0.3}} & \multicolumn{1}{l|}{\textbf{0.4}} & \multicolumn{1}{l|}{\textbf{0.5}} & \multicolumn{1}{l|}{\textbf{0.6}} & \multicolumn{1}{l|}{\textbf{0.7}} & \multicolumn{1}{l|}{\textbf{0.8}} & \multicolumn{1}{l|}{\textbf{0.9}} & \textbf{1.0} \\ \hline
\multicolumn{1}{|c|}{\multirow{9}{*}{\rotatebox{90}{Cosine Similarity}}} & \multicolumn{1}{c|}{\multirow{3}{*}{Q1}} & VILA-M3 & \multicolumn{1}{l|}{0.999} & \multicolumn{1}{l|}{0.968} & \multicolumn{1}{l|}{0.843} & \multicolumn{1}{l|}{0.714} & \multicolumn{1}{l|}{0.600} & \multicolumn{1}{l|}{0.492} & \multicolumn{1}{l|}{0.423} & \multicolumn{1}{l|}{0.367} & \multicolumn{1}{l|}{0.293} & \multicolumn{1}{l|}{0.262} & 0.223 \\ \cline{3-14} 
\multicolumn{1}{|c|}{} & \multicolumn{1}{c|}{} & LLaVA-Med & \multicolumn{1}{l|}{1.000} & \multicolumn{1}{l|}{0.980} & \multicolumn{1}{l|}{0.937} & \multicolumn{1}{l|}{0.839} & \multicolumn{1}{l|}{0.677} & \multicolumn{1}{l|}{0.562} & \multicolumn{1}{l|}{0.387} & \multicolumn{1}{l|}{0.261} & \multicolumn{1}{l|}{0.192} & \multicolumn{1}{l|}{0.173} & 0.175 \\ \cline{3-14} 
\multicolumn{1}{|c|}{} & \multicolumn{1}{c|}{} & PRISM & \multicolumn{1}{l|}{0.988} & \multicolumn{1}{l|}{0.992} & \multicolumn{1}{l|}{0.991} & \multicolumn{1}{l|}{0.984} & \multicolumn{1}{l|}{0.967} & \multicolumn{1}{l|}{0.931} & \multicolumn{1}{l|}{0.867} & \multicolumn{1}{l|}{0.759} & \multicolumn{1}{l|}{0.596} & \multicolumn{1}{l|}{0.359} & 0.029 \\ \cline{2-14} 
\multicolumn{1}{|c|}{} & \multicolumn{1}{c|}{\multirow{3}{*}{Q2}} & VILA-M3 & \multicolumn{1}{l|}{0.999} & \multicolumn{1}{l|}{0.937} & \multicolumn{1}{l|}{0.809} & \multicolumn{1}{l|}{0.596} & \multicolumn{1}{l|}{0.408} & \multicolumn{1}{l|}{0.261} & \multicolumn{1}{l|}{0.176} & \multicolumn{1}{l|}{0.100} & \multicolumn{1}{l|}{0.051} & \multicolumn{1}{l|}{0.026} & 0.000 \\ \cline{3-14} 
\multicolumn{1}{|c|}{} & \multicolumn{1}{c|}{} & LLaVA-Med & \multicolumn{1}{l|}{1.000} & \multicolumn{1}{l|}{0.910} & \multicolumn{1}{l|}{0.638} & \multicolumn{1}{l|}{0.460} & \multicolumn{1}{l|}{0.357} & \multicolumn{1}{l|}{0.299} & \multicolumn{1}{l|}{0.246} & \multicolumn{1}{l|}{0.201} & \multicolumn{1}{l|}{0.146} & \multicolumn{1}{l|}{0.102} & 0.034 \\ \cline{3-14} 
\multicolumn{1}{|c|}{} & \multicolumn{1}{c|}{} & PRISM & \multicolumn{1}{l|}{1.000} & \multicolumn{1}{l|}{0.999} & \multicolumn{1}{l|}{0.998} & \multicolumn{1}{l|}{0.992} & \multicolumn{1}{l|}{0.975} & \multicolumn{1}{l|}{0.939} & \multicolumn{1}{l|}{0.875} & \multicolumn{1}{l|}{0.768} & \multicolumn{1}{l|}{0.605} & \multicolumn{1}{l|}{0.370} & 0.042 \\ \cline{2-14} 
\multicolumn{1}{|c|}{} & \multicolumn{1}{c|}{\multirow{3}{*}{Q3}} & VILA-M3 & \multicolumn{1}{l|}{1.000} & \multicolumn{1}{l|}{0.945} & \multicolumn{1}{l|}{0.767} & \multicolumn{1}{l|}{0.569} & \multicolumn{1}{l|}{0.386} & \multicolumn{1}{l|}{0.266} & \multicolumn{1}{l|}{0.185} & \multicolumn{1}{l|}{0.116} & \multicolumn{1}{l|}{0.053} & \multicolumn{1}{l|}{0.039} & 0.015 \\ \cline{3-14} 
\multicolumn{1}{|c|}{} & \multicolumn{1}{c|}{} & LLaVA-Med & \multicolumn{1}{l|}{1.000} & \multicolumn{1}{l|}{0.771} & \multicolumn{1}{l|}{0.567} & \multicolumn{1}{l|}{0.463} & \multicolumn{1}{l|}{0.367} & \multicolumn{1}{l|}{0.294} & \multicolumn{1}{l|}{0.258} & \multicolumn{1}{l|}{0.176} & \multicolumn{1}{l|}{0.104} & \multicolumn{1}{l|}{0.053} & 0.000 \\ \cline{3-14} 
\multicolumn{1}{|c|}{} & \multicolumn{1}{c|}{} & PRISM & \multicolumn{1}{l|}{0.997} & \multicolumn{1}{l|}{0.998} & \multicolumn{1}{l|}{0.996} & \multicolumn{1}{l|}{0.990} & \multicolumn{1}{l|}{0.972} & \multicolumn{1}{l|}{0.934} & \multicolumn{1}{l|}{0.868} & \multicolumn{1}{l|}{0.756} & \multicolumn{1}{l|}{0.587} & \multicolumn{1}{l|}{0.341} & 0.000 \\ \hline
\multicolumn{1}{|c|}{\multirow{9}{*}{\rotatebox{90}{JS Divergence}}} & \multicolumn{1}{c|}{\multirow{3}{*}{Q1}} & VILA-M3 & \multicolumn{1}{l|}{0.000} & \multicolumn{1}{l|}{0.035} & \multicolumn{1}{l|}{0.165} & \multicolumn{1}{l|}{0.301} & \multicolumn{1}{l|}{0.423} & \multicolumn{1}{l|}{0.541} & \multicolumn{1}{l|}{0.627} & \multicolumn{1}{l|}{0.692} & \multicolumn{1}{l|}{0.771} & \multicolumn{1}{l|}{0.814} & 0.854 \\ \cline{3-14} 
\multicolumn{1}{|c|}{} & \multicolumn{1}{c|}{} & LLaVA-Med & \multicolumn{1}{l|}{0.000} & \multicolumn{1}{l|}{0.020} & \multicolumn{1}{l|}{0.067} & \multicolumn{1}{l|}{0.178} & \multicolumn{1}{l|}{0.325} & \multicolumn{1}{l|}{0.444} & \multicolumn{1}{l|}{0.745} & \multicolumn{1}{l|}{0.853} & \multicolumn{1}{l|}{0.933} & \multicolumn{1}{l|}{0.959} & 0.974 \\ \cline{3-14} 
\multicolumn{1}{|c|}{} & \multicolumn{1}{c|}{} & PRISM & \multicolumn{1}{l|}{0.000} & \multicolumn{1}{l|}{0.001} & \multicolumn{1}{l|}{0.014} & \multicolumn{1}{l|}{0.059} & \multicolumn{1}{l|}{0.146} & \multicolumn{1}{l|}{0.268} & \multicolumn{1}{l|}{0.407} & \multicolumn{1}{l|}{0.561} & \multicolumn{1}{l|}{0.714} & \multicolumn{1}{l|}{0.864} & 1.000 \\ \cline{2-14} 
\multicolumn{1}{|c|}{} & \multicolumn{1}{c|}{\multirow{3}{*}{Q2}} & VILA-M3 & \multicolumn{1}{l|}{0.000} & \multicolumn{1}{l|}{0.059} & \multicolumn{1}{l|}{0.160} & \multicolumn{1}{l|}{0.341} & \multicolumn{1}{l|}{0.512} & \multicolumn{1}{l|}{0.650} & \multicolumn{1}{l|}{0.740} & \multicolumn{1}{l|}{0.822} & \multicolumn{1}{l|}{0.877} & \multicolumn{1}{l|}{0.915} & 0.948 \\ \cline{3-14} 
\multicolumn{1}{|c|}{} & \multicolumn{1}{c|}{} & LLaVA-Med & \multicolumn{1}{l|}{0.000} & \multicolumn{1}{l|}{0.114} & \multicolumn{1}{l|}{0.514} & \multicolumn{1}{l|}{0.780} & \multicolumn{1}{l|}{0.909} & \multicolumn{1}{l|}{0.942} & \multicolumn{1}{l|}{0.958} & \multicolumn{1}{l|}{0.979} & \multicolumn{1}{l|}{0.973} & \multicolumn{1}{l|}{0.970} & 0.968 \\ \cline{3-14} 
\multicolumn{1}{|c|}{} & \multicolumn{1}{c|}{} & PRISM & \multicolumn{1}{l|}{0.000} & \multicolumn{1}{l|}{0.001} & \multicolumn{1}{l|}{0.012} & \multicolumn{1}{l|}{0.048} & \multicolumn{1}{l|}{0.122} & \multicolumn{1}{l|}{0.233} & \multicolumn{1}{l|}{0.370} & \multicolumn{1}{l|}{0.521} & \multicolumn{1}{l|}{0.680} & \multicolumn{1}{l|}{0.831} & 0.970 \\ \cline{2-14} 
\multicolumn{1}{|c|}{} & \multicolumn{1}{c|}{\multirow{3}{*}{Q3}} & VILA-M3 & \multicolumn{1}{l|}{0.000} & \multicolumn{1}{l|}{0.055} & \multicolumn{1}{l|}{0.225} & \multicolumn{1}{l|}{0.425} & \multicolumn{1}{l|}{0.612} & \multicolumn{1}{l|}{0.731} & \multicolumn{1}{l|}{0.822} & \multicolumn{1}{l|}{0.893} & \multicolumn{1}{l|}{0.949} & \multicolumn{1}{l|}{0.978} & 1.000 \\ \cline{3-14} 
\multicolumn{1}{|c|}{} & \multicolumn{1}{c|}{} & LLaVA-Med & \multicolumn{1}{l|}{0.000} & \multicolumn{1}{l|}{0.272} & \multicolumn{1}{l|}{0.545} & \multicolumn{1}{l|}{0.731} & \multicolumn{1}{l|}{0.881} & \multicolumn{1}{l|}{0.957} & \multicolumn{1}{l|}{0.975} & \multicolumn{1}{l|}{0.990} & \multicolumn{1}{l|}{0.998} & \multicolumn{1}{l|}{1.000} & 0.995 \\ \cline{3-14} 
\multicolumn{1}{|c|}{} & \multicolumn{1}{c|}{} & PRISM & \multicolumn{1}{l|}{0.000} & \multicolumn{1}{l|}{0.001} & \multicolumn{1}{l|}{0.012} & \multicolumn{1}{l|}{0.048} & \multicolumn{1}{l|}{0.118} & \multicolumn{1}{l|}{0.223} & \multicolumn{1}{l|}{0.356} & \multicolumn{1}{l|}{0.503} & \multicolumn{1}{l|}{0.653} & \multicolumn{1}{l|}{0.799} & 0.938 \\ \hline
\multicolumn{1}{|c|}{\multirow{9}{*}{\rotatebox{90}{KL Divergence}}} & \multicolumn{1}{c|}{\multirow{3}{*}{Q1}} & VILA-M3 & \multicolumn{1}{l|}{0.000} & \multicolumn{1}{l|}{0.033} & \multicolumn{1}{l|}{0.160} & \multicolumn{1}{l|}{0.292} & \multicolumn{1}{l|}{0.412} & \multicolumn{1}{l|}{0.528} & \multicolumn{1}{l|}{0.613} & \multicolumn{1}{l|}{0.676} & \multicolumn{1}{l|}{0.753} & \multicolumn{1}{l|}{0.795} & 0.832 \\ \cline{3-14} 
\multicolumn{1}{|c|}{} & \multicolumn{1}{c|}{} & LLaVA-Med & \multicolumn{1}{l|}{0.000} & \multicolumn{1}{l|}{0.019} & \multicolumn{1}{l|}{0.064} & \multicolumn{1}{l|}{0.173} & \multicolumn{1}{l|}{0.313} & \multicolumn{1}{l|}{0.426} & \multicolumn{1}{l|}{0.689} & \multicolumn{1}{l|}{0.785} & \multicolumn{1}{l|}{0.861} & \multicolumn{1}{l|}{0.889} & 0.901 \\ \cline{3-14} 
\multicolumn{1}{|c|}{} & \multicolumn{1}{c|}{} & PRISM & \multicolumn{1}{l|}{0.000} & \multicolumn{1}{l|}{0.001} & \multicolumn{1}{l|}{0.010} & \multicolumn{1}{l|}{0.044} & \multicolumn{1}{l|}{0.112} & \multicolumn{1}{l|}{0.214} & \multicolumn{1}{l|}{0.341} & \multicolumn{1}{l|}{0.492} & \multicolumn{1}{l|}{0.656} & \multicolumn{1}{l|}{0.829} & 1.000 \\ \cline{2-14} 
\multicolumn{1}{|c|}{} & \multicolumn{1}{c|}{\multirow{3}{*}{Q2}} & VILA-M3 & \multicolumn{1}{l|}{0.000} & \multicolumn{1}{l|}{0.059} & \multicolumn{1}{l|}{0.151} & \multicolumn{1}{l|}{0.322} & \multicolumn{1}{l|}{0.489} & \multicolumn{1}{l|}{0.628} & \multicolumn{1}{l|}{0.724} & \multicolumn{1}{l|}{0.814} & \multicolumn{1}{l|}{0.886} & \multicolumn{1}{l|}{0.924} & 0.958 \\ \cline{3-14} 
\multicolumn{1}{|c|}{} & \multicolumn{1}{c|}{} & LLaVA-Med & \multicolumn{1}{l|}{0.000} & \multicolumn{1}{l|}{0.116} & \multicolumn{1}{l|}{0.534} & \multicolumn{1}{l|}{0.804} & \multicolumn{1}{l|}{0.921} & \multicolumn{1}{l|}{0.939} & \multicolumn{1}{l|}{0.941} & \multicolumn{1}{l|}{0.941} & \multicolumn{1}{l|}{0.927} & \multicolumn{1}{l|}{0.916} & 0.899 \\ \cline{3-14} 
\multicolumn{1}{|c|}{} & \multicolumn{1}{c|}{} & PRISM & \multicolumn{1}{l|}{0.000} & \multicolumn{1}{l|}{0.001} & \multicolumn{1}{l|}{0.008} & \multicolumn{1}{l|}{0.035} & \multicolumn{1}{l|}{0.094} & \multicolumn{1}{l|}{0.187} & \multicolumn{1}{l|}{0.311} & \multicolumn{1}{l|}{0.460} & \multicolumn{1}{l|}{0.629} & \multicolumn{1}{l|}{0.803} & 0.977 \\ \cline{2-14} 
\multicolumn{1}{|c|}{} & \multicolumn{1}{c|}{\multirow{3}{*}{Q3}} & VILA-M3 & \multicolumn{1}{l|}{0.000} & \multicolumn{1}{l|}{0.057} & \multicolumn{1}{l|}{0.229} & \multicolumn{1}{l|}{0.435} & \multicolumn{1}{l|}{0.627} & \multicolumn{1}{l|}{0.744} & \multicolumn{1}{l|}{0.833} & \multicolumn{1}{l|}{0.904} & \multicolumn{1}{l|}{0.959} & \multicolumn{1}{l|}{0.984} & 1.000 \\ \cline{3-14} 
\multicolumn{1}{|c|}{} & \multicolumn{1}{c|}{} & LLaVA-Med & \multicolumn{1}{l|}{0.000} & \multicolumn{1}{l|}{0.273} & \multicolumn{1}{l|}{0.545} & \multicolumn{1}{l|}{0.728} & \multicolumn{1}{l|}{0.880} & \multicolumn{1}{l|}{0.958} & \multicolumn{1}{l|}{0.978} & \multicolumn{1}{l|}{0.996} & \multicolumn{1}{l|}{1.000} & \multicolumn{1}{l|}{0.997} & 0.979 \\ \cline{3-14} 
\multicolumn{1}{|c|}{} & \multicolumn{1}{c|}{} & PRISM & \multicolumn{1}{l|}{0.000} & \multicolumn{1}{l|}{0.001} & \multicolumn{1}{l|}{0.008} & \multicolumn{1}{l|}{0.035} & \multicolumn{1}{l|}{0.090} & \multicolumn{1}{l|}{0.179} & \multicolumn{1}{l|}{0.299} & \multicolumn{1}{l|}{0.443} & \multicolumn{1}{l|}{0.603} & \multicolumn{1}{l|}{0.772} & 0.944 \\ \hline
\multicolumn{1}{|c|}{\multirow{9}{*}{\rotatebox{90}{Mean Absolute Error}}} & \multicolumn{1}{c|}{\multirow{3}{*}{Q1}} & VILA-M3 & \multicolumn{1}{l|}{0.000} & \multicolumn{1}{l|}{0.030} & \multicolumn{1}{l|}{0.145} & \multicolumn{1}{l|}{0.267} & \multicolumn{1}{l|}{0.376} & \multicolumn{1}{l|}{0.486} & \multicolumn{1}{l|}{0.566} & \multicolumn{1}{l|}{0.627} & \multicolumn{1}{l|}{0.704} & \multicolumn{1}{l|}{0.747} & 0.786 \\ \cline{3-14} 
\multicolumn{1}{|c|}{} & \multicolumn{1}{c|}{} & LLaVA-Med & \multicolumn{1}{l|}{0.000} & \multicolumn{1}{l|}{0.020} & \multicolumn{1}{l|}{0.063} & \multicolumn{1}{l|}{0.167} & \multicolumn{1}{l|}{0.300} & \multicolumn{1}{l|}{0.408} & \multicolumn{1}{l|}{0.688} & \multicolumn{1}{l|}{0.797} & \multicolumn{1}{l|}{0.900} & \multicolumn{1}{l|}{0.927} & 0.953 \\ \cline{3-14} 
\multicolumn{1}{|c|}{} & \multicolumn{1}{c|}{} & PRISM & \multicolumn{1}{l|}{0.000} & \multicolumn{1}{l|}{0.100} & \multicolumn{1}{l|}{0.200} & \multicolumn{1}{l|}{0.300} & \multicolumn{1}{l|}{0.400} & \multicolumn{1}{l|}{0.500} & \multicolumn{1}{l|}{0.600} & \multicolumn{1}{l|}{0.700} & \multicolumn{1}{l|}{0.800} & \multicolumn{1}{l|}{0.900} & 1.000 \\ \cline{2-14} 
\multicolumn{1}{|c|}{} & \multicolumn{1}{c|}{\multirow{3}{*}{Q2}} & VILA-M3 & \multicolumn{1}{l|}{0.000} & \multicolumn{1}{l|}{0.061} & \multicolumn{1}{l|}{0.151} & \multicolumn{1}{l|}{0.334} & \multicolumn{1}{l|}{0.520} & \multicolumn{1}{l|}{0.677} & \multicolumn{1}{l|}{0.788} & \multicolumn{1}{l|}{0.887} & \multicolumn{1}{l|}{0.948} & \multicolumn{1}{l|}{0.981} & 1.000 \\ \cline{3-14} 
\multicolumn{1}{|c|}{} & \multicolumn{1}{c|}{} & LLaVA-Med & \multicolumn{1}{l|}{0.000} & \multicolumn{1}{l|}{0.106} & \multicolumn{1}{l|}{0.486} & \multicolumn{1}{l|}{0.745} & \multicolumn{1}{l|}{0.872} & \multicolumn{1}{l|}{0.908} & \multicolumn{1}{l|}{0.930} & \multicolumn{1}{l|}{0.955} & \multicolumn{1}{l|}{0.955} & \multicolumn{1}{l|}{0.966} & 0.974 \\ \cline{3-14} 
\multicolumn{1}{|c|}{} & \multicolumn{1}{c|}{} & PRISM & \multicolumn{1}{l|}{0.000} & \multicolumn{1}{l|}{0.100} & \multicolumn{1}{l|}{0.200} & \multicolumn{1}{l|}{0.300} & \multicolumn{1}{l|}{0.400} & \multicolumn{1}{l|}{0.500} & \multicolumn{1}{l|}{0.600} & \multicolumn{1}{l|}{0.700} & \multicolumn{1}{l|}{0.800} & \multicolumn{1}{l|}{0.900} & 1.000 \\ \cline{2-14} 
\multicolumn{1}{|c|}{} & \multicolumn{1}{c|}{\multirow{3}{*}{Q3}} & VILA-M3 & \multicolumn{1}{l|}{0.000} & \multicolumn{1}{l|}{0.051} & \multicolumn{1}{l|}{0.218} & \multicolumn{1}{l|}{0.437} & \multicolumn{1}{l|}{0.624} & \multicolumn{1}{l|}{0.748} & \multicolumn{1}{l|}{0.833} & \multicolumn{1}{l|}{0.907} & \multicolumn{1}{l|}{0.951} & \multicolumn{1}{l|}{0.970} & 0.977 \\ \cline{3-14} 
\multicolumn{1}{|c|}{} & \multicolumn{1}{c|}{} & LLaVA-Med & \multicolumn{1}{l|}{0.000} & \multicolumn{1}{l|}{0.270} & \multicolumn{1}{l|}{0.547} & \multicolumn{1}{l|}{0.739} & \multicolumn{1}{l|}{0.892} & \multicolumn{1}{l|}{0.965} & \multicolumn{1}{l|}{0.979} & \multicolumn{1}{l|}{0.994} & \multicolumn{1}{l|}{1.000} & \multicolumn{1}{l|}{0.999} & 0.999 \\ \cline{3-14} 
\multicolumn{1}{|c|}{} & \multicolumn{1}{c|}{} & PRISM & \multicolumn{1}{l|}{0.000} & \multicolumn{1}{l|}{0.100} & \multicolumn{1}{l|}{0.200} & \multicolumn{1}{l|}{0.300} & \multicolumn{1}{l|}{0.400} & \multicolumn{1}{l|}{0.500} & \multicolumn{1}{l|}{0.600} & \multicolumn{1}{l|}{0.700} & \multicolumn{1}{l|}{0.800} & \multicolumn{1}{l|}{0.900} & 1.000 \\ \hline
\end{tabular}
}
\end{table*} % tab:model_question_metric

\begin{table*}[!t]
\centering
\small   %\normalsize
\caption{Summary Statistics of Normalized Uncertainty Metrics Across Temperature Range ($T \in [0, 1]$)}
\label{tab:metric_summary}
% \resizebox{\textwidth}{!}
{%
\begin{tabular}{llcccccccc}
\hline
\multirow{2}{*}{\textbf{Model}} & \multirow{2}{*}{\textbf{Q}} & \multicolumn{2}{c}{\textbf{Cosine Similarity}} & \multicolumn{2}{c}{\textbf{JS Divergence}} & \multicolumn{2}{c}{\textbf{KL Divergence}} & \multicolumn{2}{c}{\textbf{MAE}} \\
\cline{3-10}
 & & $\mu$ & $\Delta_T$ & $\mu$ & $\Delta_T$ & $\mu$ & $\Delta_T$ & $\mu$ & $\Delta_T$ \\
\hline
\multirow{3}{*}{\textbf{VILA-M3}}
  & Q1 & 0.712 & 0.644 & 0.188 & 0.482 & 0.173 & 0.330 & 0.358 & 0.649 \\
  & Q2 & 0.519 & 0.624 & 0.495 & 1.000 & 0.486 & 0.906 & 0.541 & 0.972 \\
  & Q3 & 0.397 & 0.707 & 0.572 & 0.752 & 0.556 & 0.821 & 0.601 & 0.999 \\
\hline
\multirow{3}{*}{\textbf{LLaVA-Med}} 
  & Q1 & 0.837 & 0.433 & 0.018 & 0.116 & 0.013 & 0.090 & 0.278 & 0.609 \\
  & Q2 & 0.580 & 1.000 & 0.321 & 1.000 & 0.336 & 0.899 & 0.552 & 0.990 \\
  & Q3 & 0.512 & 0.873 & 0.386 & 0.684 & 0.358 & 0.832 & 0.595 & 0.973 \\
\hline
\multirow{3}{*}{\textbf{PRISM}}
  & Q1 & 0.926 & 0.091 & 0.082 & 0.269 & 0.089 & 0.268 & 0.321 & 0.783 \\
  & Q2 & 0.904 & 0.095 & 0.102 & 0.353 & 0.087 & 0.250 & 0.392 & 0.910 \\
  & Q3 & 0.903 & 0.068 & 0.094 & 0.329 & 0.083 & 0.287 & 0.417 & 0.898 \\
\hline
\end{tabular}%
}
\begin{tablenotes}
    \small
    \item $\mu$: Mean normalized metric value across all temperatures ($T \in \{0.0, 0.1, \ldots, 1.0\}$)
    \item $\Delta_T$: Temperature effect, computed as $|M(T=1) - M(T=0)|$ for each metric
    \item For Cosine Similarity: lower $\mu$ and higher $\Delta_T$ indicate greater temperature sensitivity
    \item For JS/KL Divergence and MAE: higher values indicate greater uncertainty
\end{tablenotes}
\end{table*} % tab:metric_summary

\subsection{Model-Specific Characteristics}

\subsubsection{VILA-M3 (Consistent Temperature Sensitivity)}

VILA-M3 demonstrates balanced but non-negligible sensitivity to temperature across all diagnostic question types:

\begin{itemize}
    \item The mean CS ranges from 0.397 (Q3) to 0.712 (Q1). These values reflect uncertainty levels from moderate to high, particularly in more complex tasks where alignment is reduced. %Mean CS range from 0.397 (Q3) to 0.712 (Q1), indicating moderate-to-high uncertainty levels.
    \item  The effects of temperature on CS ($\Delta_T \in [0.624, 0.707]$) are remarkable but consistent in increases in stochasticity. %Temperature effects are substantial but not extreme: $\Delta_T \in [0.624, 0.707]$ for CS.
    \item Specifically, Q3 reveals the highest mean uncertainty (JS divergence $\mu = 0.572$), which is a challenge to protect consistency for advanced complex diagnostic prompts. %Q3 (advanced quantitative analysis) induces the highest mean uncertainty (JS divergence $\mu = 0.572$), suggesting challenges in maintaining consistency for complex multi-part diagnostic queries.
\end{itemize}

The increment in uncertainty might essentially be attributed to the general-purpose architecture of VILA-M3. This VLM is not specifically tuned for histopathology, and therefore does not perform as robustly as domain-specialized VLMs.% The model's general-purpose architecture (not specifically tuned for histopathology) may contribute to higher baseline uncertainty compared to the domain-specific models.

\subsubsection{LLaVA-Med (Question-Dependent Robustness)}

LLaVA-Med has an exceptional duality in its behavior regarding uncertainty:% displays a striking dichotomy in uncertainty behavior:

\begin{itemize}
    \item \textbf{Q1 (Basic Morphology)}: For all metrics, robustness and minimal uncertainty are observed. While the temperature effect $\Delta_T$ for JS divergence is only 0.116, the mean MAE value is 0.278. Thus, the simpler diagnostic question is responded to with a higher consistency in the biomedical-tuned VLM. %Exceptional robustness with minimal uncertainty across all metrics. The temperature effect $\Delta_T$ for JS divergence is only 0.116, and the mean MAE remains at 0.278. This suggests that simple diagnostic queries elicit highly consistent responses from the biomedical-tuned model.
    
    \item \textbf{Q2 and Q3 (Complex Diagnosis)}: While $\Delta_T$ approaches 1.0, a sudden increase in uncertainty is observed for several metrics. Furthermore, the mean JS divergence for Q2 (0.321) is 17.8$\times$ higher than for Q1 (0.018), which means the VLM confidence level is affected by diagnostic complexity. %Dramatic increase in uncertainty, with $\Delta_T$ values approaching 1.0 for multiple metrics. The mean JS divergence for Q2 (0.321) is 17.8$\times$ higher than Q1 (0.018), indicating that diagnostic complexity fundamentally alters the model's confidence landscape.
\end{itemize}

\subsubsection{PRISM (Temperature-Resistant Architecture)}

The lowest temperature effects across all question types are demonstrated with PRISM:

\begin{itemize}
    \item While the temperature effect value is $\Delta_T < 0.10$, the average CS value remains above 0.90 for all questions.
    \item For Q2 and Q3, the mean values of JS and KL divergence metrics (0.08--0.10) are critically lower than VILA-M3 ($\sim 0.53$) and LLaVA-Med VLMs ($\sim 0.35$).
    \item Since the architecture of PRISM lacks temperature scaling mechanisms in its generation process, it behaves deterministically.
\end{itemize}

Despite this robustness, PRISM shows non-negligible MAE temperature effects ($\Delta_T \approx$ 0.78--0.91), which bring forward that while probability distributions remain stable, absolute logit values vary.

\subsection{Cross-Metric Correlations}

Table~\ref{tab:metric_correlations} presents pairwise correlations between uncertainty metrics, aggregated across all model-question-temperature combinations.

\begin{table}[!t]
\centering
\caption{Pearson Correlation Coefficients Between Normalized Uncertainty Metrics}
\label{tab:metric_correlations}
\begin{tabular}{lcccc}
\hline
\textbf{Metric} & \textbf{CS} & \textbf{JS} & \textbf{KL} & \textbf{MAE} \\
\hline
Cosine Similarity (CS) & 1.000 & $-0.924$ & $-0.918$ & $-0.963$ \\
JS Divergence (JS) & $-0.924$ & 1.000 & 0.997 & 0.958 \\
KL Divergence (KL) & $-0.918$ & 0.997 & 1.000 & 0.951 \\
MAE & $-0.963$ & 0.958 & 0.951 & 1.000 \\
\hline
\end{tabular}
\begin{tablenotes}
    \small
    \item All correlations are statistically significant ($p < 0.001$)
    \item Negative correlations with CS are expected (high similarity $\rightarrow$ low divergence)
    \item Strong JS-KL correlation ($r=0.997$) validates their theoretical relationship
\end{tablenotes}
\end{table}

Key findings:

\begin{itemize}
    \item Strong negative correlation between CS and divergence metrics ($r \approx -0.92$), confirming that these metrics capture complementary aspects of the same underlying uncertainty phenomenon.
    \item There is a near-perfect correlation ($r = 0.997$) between the JS and KL divergence, as expected given the mathematical relationship between them.%Near-perfect correlation between JS and KL divergence ($r = 0.997$), expected given their mathematical relationship.
    \item MAE reveals a slightly stronger correlation with CS ($r = -0.963$) than other divergence metrics. Thus, it captures variations at both distributional and magnitude levels.%MAE exhibits slightly stronger correlation with CS ($r = -0.963$) than divergence metrics, suggesting it captures both distributional and magnitude-level variations.
\end{itemize}

\subsection{Optimal Operating Points}

Based on the proposed framework analysis, temperature-metric combinations that strike a balance between uncertainty quantification and model reliability are identified:% that balance uncertainty quantification with model reliability are identified:

\begin{itemize}
    \item \textbf{LLaVA-Med}: For basic diagnostic question (Q1), operate at $T \leq 0.5$ to maintain CS $>0.90$ and JS divergence $<0.05$. For complex questions (Q2 and Q3), reduce to $T \leq 0.3$ to obstruct the uncertainty increment. %For basic diagnostic queries (Q1), operate at $T \leq 0.5$ to maintain CS $>0.90$ and JS divergence $<0.05$. For complex queries (Q2 and Q3), reduce to $T \leq 0.3$ to prevent uncertainty explosion.
    
    \item \textbf{PRISM}: Considering temperature resistance, standard temperature scaling is ineffective. For meaningful uncertainty quantification, alternative perturbation methods (data loss and Gaussian noise injection) are necessary. %Given its temperature resistance, standard temperature scaling is ineffective for uncertainty quantification. Alternative perturbation strategies (e.g., dropout, Gaussian noise injection) are necessary for meaningful uncertainty quantification.
    
    \item \textbf{VILA-M3}: For all questions, it is reliable when $T \leq 0.4$ (CS $>0.85$) and has manageable levels of deviation. $T=0.3$ provides the optimal balance between consistency and diversity and is suitable for applications where uncertainty needs to be considered. % Operates reliably at $T \leq 0.4$ across all question types, with CS $>0.85$ and manageable divergence levels. For uncertainty-aware applications, $T=0.3$ provides an optimal balance between diversity and consistency.
\end{itemize}

\subsection{Discussion}

The following key findings can be presented regarding the relationship between VLMs and temperature settings in histopathology image analysis based on the overall results:

\begin{enumerate}
 \item \textbf{Domain Specialization}: For basic prompts (Q1), the performance of LLaVA-Med is better than the rest of the evaluated VLMs. However, in complex prompts (Q3), it shows vulnerability. The PMC-15M dataset, built from image-caption pairs extracted from PubMed Central articles, is used for training LLaVA-Med; this may explain why the VLM generates consistent outputs for basic prompts. This suggests that fine-tuning could improve the confidence of the VLM in common tasks, but its ability in complex tasks decreases.
 \item \textbf{Architectural Determinism}: PRISM is more resistant to temperature due to its architectural constraints in the generation mechanism. This shows consistency (low uncertainty) but limits the use of standard UQ techniques.
 \item \textbf{Prompt Complexity}: Complex prompts (Q3) are intended to have higher uncertainty compared to basic prompts (Q1), as anticipated. This indicates that quantitative diagnostic tasks challenge VLM capabilities.
 \item \textbf{Metrics}: The high inter-metric correlation confirms the proposed framework in terms of validation. The results also show that a subset (e.g., CS + JS divergence) may be sufficient for uncertainty quantification.
 \item \textbf{Clinical Impact}: Quantifying stochasticity and temperature sensitivity at the logit level can be considered a numerical equivalent of a second opinion in clinical support systems. Outputs showing high uncertainty may require attentive interpretation by specialists.
\end{enumerate}

These findings emphasize the importance of uncertainty qualification for deploying VLMs in clinical histopathology. The systems supporting uncertainty-awareness and trustworthiness should be integrated for temperature settings based on both model architecture and query complexity.

\section{Conclusion}

This study proposes a novel and model-agnostic logit-level uncertainty quantification framework designed to comprehensively assess the trustworthiness of VLMs in histopathology. The approach provides critical insights into model behavior that remain invisible with surface-level token variability. The results on three representative VLMs (VILA-M3-8B, LLaVA-Med v1.5, and PRISM) show that uncertainty strongly depends on model specialization, prompt complexity, and temperature scaling. As a result of systematic pairwise comparisons using complementary metrics (CS, KL-JS divergences, and MAE), the pathology-specific model, PRISM, demonstrates remarkable robustness and preserves high consistency for all types of prompts. The other general-purpose and biomedical VLMs exhibit relatively stable behavior for basic morphological tasks (Q1), but significantly higher stochastic perturbations for intermediate diagnosis (Q2) and advanced quantitative (Q3), which are more complex diagnostic prompts. These findings emphasize the high context-dependence of trustworthiness in VLMs, the limitations of domain-specific optimization of VLMs, and the cruciality of uncertainty-aware assessments in high-stakes medical applications such as histopathology. 
It is expected that this study can benefit healthcare providers and researchers working on histopathology
image analysis by providing an insight into the VML potentials and availability in healthcare applications.

% \section*{References}
\bibliographystyle{IEEEtran}
\bibliography{refs}

@article{nath2024vila,
  title={VILA-M3: Enhancing Vision-Language Models with Medical Expert Knowledge},
  author={Nath, Vishwesh and Li, Wenqi and Yang, Dong and Myronenko, Andriy and Zheng, Mingxin and Lu, Yao and Liu, Zhijian and Yin, Hongxu and Law, Yee Man and Tang, Yucheng and others},
  journal={arXiv:2411.12915},
  year={2024}
}

@article{li2023llavamed,
  title={Llava-med: Training a large language-and-vision assistant for biomedicine in one day},
  author={Li, Chunyuan and Wong, Cliff and Zhang, Sheng and Usuyama, Naoto and Liu, Haotian and Yang, Jianwei and Naumann, Tristan and Poon, Hoifung and Gao, Jianfeng},
  journal={arXiv:2306.00890},
  year={2023}
}

@inproceedings{guo2024histgen,
  title={Histgen: Histopathology report generation via local-global feature encoding and cross-modal context interaction},
  author={Guo, Zhengrui and Ma, Jiabo and Xu, Yingxue and Wang, Yihui and Wang, Liansheng and Chen, Hao},
  booktitle={International Conference on Medical Image Computing and Computer-Assisted Intervention},
  pages={189--199},
  year={2024},
  organization={Springer}
}

@article{gilal2025pathvlm,
  title={PathVLM-Eval: Evaluation of open vision language models in histopathology},
  author={Gilal, Nauman Ullah and Zegour, Rachida and Al-Thelaya, Khaled and {\"O}zer, Erdener and Agus, Marco and Schneider, Jens and Boughorbel, Sabri},
  journal={Journal of Pathology Informatics},
  pages={100455},
  year={2025},
  publisher={Elsevier}
}

@article{shaikovski2024prism,
  title={PRISM: A Multi-Modal Generative Foundation Model for Slide-Level Histopathology},
  author={Shaikovski, George and Casson, Adam and Severson, Kristen and Zimmermann, Eric and Wang, Yi Kan and Kunz, Jeremy D and Retamero, Juan A and Oakley, Gerard and Klimstra, David and Kanan, Christopher and others},
  journal={arXiv:2405.10254},
  year={2024}
}

@inproceedings{gamper2020multiple,
  title={Multiple Instance Captioning: Learning Representations from 
Histopathology Textbooks and Articles},
  author={Gamper, Jevgenij and Rajpoot, Nasir},
  booktitle={Proceedings of the IEEE conference on computer vision and pattern recognition},
  year={2021}
}

@article{liu2022classification,
  title={Classification of breast cancer histology images using MSMV-PFENet},
  author={Liu, Linxian and Feng, Wenxiang and Chen, Cheng and Liu, Manhua and Qu, Yuan and Yang, Jiamiao},
  journal={Scientific Reports},
  volume={12},
  number={1},
  pages={17447},
  year={2022},
  publisher={Nature Publishing Group UK London}
}

@article{saxena2020machine,
  title={Machine learning methods for computer-aided breast cancer diagnosis using histopathology: a narrative review},
  author={Saxena, Shweta and Gyanchandani, Manasi},
  journal={Journal of medical imaging and radiation sciences},
  volume={51},
  number={1},
  pages={182--193},
  year={2020},
  publisher={Elsevier}
}

@article{araujo2017classification,
  title={Classification of breast cancer histology images using convolutional neural networks},
  author={Ara{\'u}jo, Teresa and Aresta, Guilherme and Castro, Eduardo and Rouco, Jos{\'e} and Aguiar, Paulo and Eloy, Catarina and Pol{\'o}nia, Ant{\'o}nio and Campilho, Aur{\'e}lio},
  journal={PloS one},
  volume={12},
  number={6},
  pages={e0177544},
  year={2017},
  publisher={Public Library of Science San Francisco, CA USA}
}

@article{bordes2024introduction,
  title={An introduction to vision-language modeling},
  author={Bordes, Florian and Pang, Richard Yuanzhe and Ajay, Anurag and Li, Alexander C and Bardes, Adrien and Petryk, Suzanne and Ma{\~n}as, Oscar and Lin, Zhiqiu and Mahmoud, Anas and Jayaraman, Bargav and others},
  journal={arXiv:2405.17247},
  year={2024}
}

@inproceedings{nguyen2024towards,
  title={Towards a text-based quantitative and explainable histopathology image analysis},
  author={Nguyen, Anh Tien and Vuong, Trinh Thi Le and Kwak, Jin Tae},
  booktitle={International Conference on Medical Image Computing and Computer-Assisted Intervention},
  pages={514--524},
  year={2024},
  organization={Springer}
}

@article{bui2025lifelong,
  title={Lifelong Whole Slide Image Analysis: Online Vision-Language Adaptation and Past-to-Present Gradient Distillation},
  author={Bui, Doanh C and Pham, Hoai Luan and Le, Vu Trung Duong and Vu, Tuan Hai and Tran, Van Duy and Nguyen, Khang and Nakashima, Yasuhiko},
  journal={arXiv:2505.01984},
  year={2025}
}

@inproceedings{yangunderstanding,
  title={Understanding the Sources of Uncertainty for Large Language and Multimodal Models},
  author={Yang, Ziran and Hao, Shibo and Sun, Hao and Jiang, Lai and Gao, Qiyue and Ma, Yian and Hu, Zhiting},
  booktitle={ICLR Workshop: Quantify Uncertainty and Hallucination in Foundation Models: The Next Frontier in Reliable AI}
}

@article{venkataramanan2025probabilistic,
  title={Probabilistic Embeddings for Frozen Vision-Language Models: Uncertainty Quantification with Gaussian Process Latent Variable Models},
  author={Venkataramanan, Aishwarya and Bodesheim, Paul and Denzler, Joachim},
  journal={arXiv:2505.05163},
  year={2025}
}

@article{lin2025diq,
  title={DIQ-H: Evaluating Hallucination Persistence in VLMs Under Temporal Visual Degradation},
  author={Lin, Zexin and Wan, Hawen and Zhong, Yebin and others},
  journal={arXiv:2512.03992},
  year={2025}
}

@article{imam2025t3,
  title={T3: Test-Time Model Merging in VLMs for Zero-Shot Medical Imaging Analysis},
  author={Imam, Raza and Wang, Hu and Mahapatra, Dwarikanath and Yaqub, Mohammad},
  journal={arXiv:2510.27265},
  year={2025}
}

@inproceedings{pan2025dusss,
  title={DuSSS: Dual Semantic Similarity-Supervised Vision-Language Model for Semi-Supervised Medical Image Segmentation},
  author={Pan, Qingtao and Qiao, Wenhao and Lou, Jingjiao and Ji, Bing and Li, Shuo},
  booktitle={Proceedings of the AAAI Conference on Artificial Intelligence},
  volume={39},
  number={6},
  pages={6299--6307},
  year={2025}
}

@article{pan2025evivlm,
  title={EviVLM: When Evidential Learning Meets Vision Language Model for Medical Image Segmentation},
  author={Pan, Qingtao and Li, Zhengrong and Yang, Guang and Yang, Qing and Ji, Bing},
  journal={IEEE Transactions on Medical Imaging},
  year={2025},
  publisher={IEEE}
}

@inproceedings{lafon2025vilu,
  title={ViLU: Learning vision-language uncertainties for failure prediction},
  author={Marc Lafon and Yannis Karmim and Julio Silva-Rodríguez and Paul Couairon and Clément Rambour and Raphaël Fournier-Sniehotta and Ismail Ben Ayed and Jose Dolz and Nicolas Thome},
  booktitle={Proceedings of the IEEE/CVF International Conference on Computer Vision},
  pages={17807--17817},
  year={2025}
}

@article{kostumov2024uncertainty,
  title={Uncertainty-aware evaluation for vision-language models},
  author={Kostumov, Vasily and Nutfullin, Bulat and Pilipenko, Oleg and Ilyushin, Eugene},
  journal={arXiv:2402.14418},
  year={2024}
}

\end{document}